\documentclass[preprint,12pt]{elsarticle}

\usepackage{amsmath}
\usepackage{amssymb}
\usepackage{tabularx}
\usepackage{booktabs}
\usepackage{subcaption}
\usepackage{multirow}
\usepackage{xcolor}
\usepackage{caption}

\begin{document}

\title{Quantifying Spectral Differences in Vehicle Kinematics Between Production Autonomous and Human-Driven Vehicles Across Driving Scenarios}

\author[1]{Peiyi Fang}
\author[1]{Xiangyu Li}
\author[1]{Yonglin Weng}
\author[1]{Ke Ma\corref{cor1}}
\ead{kema@hkust-gz.edu.cn}
\affiliation[1]{
    organization={The Hong Kong University of Science and Technology (Guangzhou)},
    city={Guangzhou},
    postcode={511453},
    country={China}
}
\cortext[cor1]{Corresponding author}

\begin{abstract}

Differences in vehicle kinematic characteristics between production autonomous vehicles (PAVs) and human-driven vehicles (HVs) have been limitedly investigated by empirical studies. Most recent studies rely on simulation-based models, while some further investigate low-level adaptive cruise control (ACC) systems in controlled experiments. These methods commonly adapt some time-domain metrics to characterize PAV-HV differences across limited driving conditions. However, current PAVs equipped with high-level autonomous driving systems generate driving behaviors in a black box using data-driven models. These fundamentally different mechanisms for generating behaviors may produce distinct kinematic characteristics in traffic.  More importantly, these time-domain metrics cannot reflect frequency-related traffic dynamics across different driving scenarios.  Thus, this study adapted a real-world PAV dataset with four PAV platforms and developed a frequency-domain framework to quantify kinematic differences between PAVs and HVs across diverse driving scenarios, including varying driving states, lighting, weather, and vehicle densities. The framework transforms kinematic signals into the frequency domain and extracts spectral features, and then compares these features between PAVs and HVs based on kernel density estimation and Wasserstein distance. The results reveal clear scenario-dependent PAV-HV spectral differences. Specifically, speed-related differences were consistently smaller during car-following than cruising, while rainy conditions consistently enlarged acceleration-related differences compared with clear conditions. These findings highlight the necessity of multi-scenario evaluations and demonstrate the value of frequency-domain analysis for characterizing PAV-HV kinematic differences under real-world conditions.

\end{abstract}


\maketitle

\section{Introduction}

Production autonomous vehicles (PAVs) are gradually being deployed on public roads alongside human-driven vehicles (HVs).
However, the differences in kinematic characteristics between these two types of vehicles remain insufficiently investigated. 
HVs rely on human perception, judgment, and control, whereas PAVs generate driving behaviors through data-driven models with largely black-box internal decision-making processes. 
The contrasting decision-making and control processes can produce distinct kinematic responses under similar traffic conditions.
Such kinematic differences may compromise the applicability of existing behavior models, trajectory prediction methods, traffic flow analyses, and safety assessment frameworks in PAV--HV traffic systems~\cite{Toledo2007,Mozaffari2022}. 
Furthermore, driving scenarios impose varying longitudinal and lateral control demands on vehicles, including speed regulation, car-following adjustment, and turning control~\cite{Zhao2020,Sun2026Scenario}. 
These demands may lead to changes in both the magnitude and pattern of PAV--HV kinematic differences.
Therefore, systematic analysis of real-world data across multiple driving scenarios is needed to quantify these differences and further understand PAV behavior.


Existing studies have commonly characterized PAV behavior through simulation-based models or experiments involving low-level adaptive cruise control (ACC) systems. 
Simulation-based studies typically represent PAV behavior through predefined behavioral or control rules~\cite{Song2023,Zheng2020}, while ACC-based studies mainly examine longitudinal car-following under limited operating conditions~\cite{Shi2021CarFollowing,Ye2023ACC}. 
These approaches cannot fully represent the behavior of current PAVs equipped with high-level automated driving systems (ADS), which operate based on broader environmental information, predictive decision-making, and multiple driving objectives.
This gap highlights the need for direct empirical analysis of current PAV behavior, yet only a few studies have examined PAV behavior based on real-world vehicle data.
These studies are typically restricted to specific vehicle platforms or a limited range of driving scenarios.
For instance, on-road testing has characterized the speed-following behavior of a Tesla Model 3 under Autopilot and Full Self-Driving modes~\cite{Duoba2024}.
Constrained by experimental scale and scenario coverage, such studies remain insufficient for systematically characterizing PAVs kinematic characteristics in complex real-world traffic environments. 
Meanwhile, other studies have leveraged large-scale public trajectory datasets containing PAV data, such as the Waymo Open Dataset~\cite{Sun2020Waymo}. 
Although these datasets enable empirical comparisons between PAVs and HVs~\cite{Wang2023Waymo,Hu2023Waymo}, limitations in vehicle-type identification, driving-scenario annotation, and trajectory continuity hinder systematic comparisons across diverse driving scenarios.

Beyond limitations in PAV data, existing studies commonly rely on task-specific time-domain metrics to characterize vehicle behavior. 
In longitudinal interaction scenarios, behavioral differences are commonly described using metrics such as speed, acceleration, spacing, and car-following stability~\cite{Talebpour2016Influence,Shi2021CarFollowing,Milanes2014}. 
Lateral driving behavior is typically characterized using trajectory features, changes in lateral position, and dynamic measures~\cite{Toledo2007,Schubert2010}. 
Such scenario-specific metric selection complicates systematic comparison across driving conditions.
Moreover, employed time-domain metrics mainly characterize kinematic variables through statistical measures such as the mean, variance, and peak values. However, signals with similar statistical values may still exhibit substantial differences in their rates of change and fluctuation patterns. This indicates that such metrics cannot fully capture the dynamic characteristics of kinematic signals. 
Fundamental traffic dynamics, including traffic oscillations, wave propagation, and disturbance propagation, are directly related to the dynamic evolution of vehicle responses~\cite{Li2010Oscillation}. 

\begin{figure}[!htb]
\centering
\includegraphics[width=0.98\textwidth]{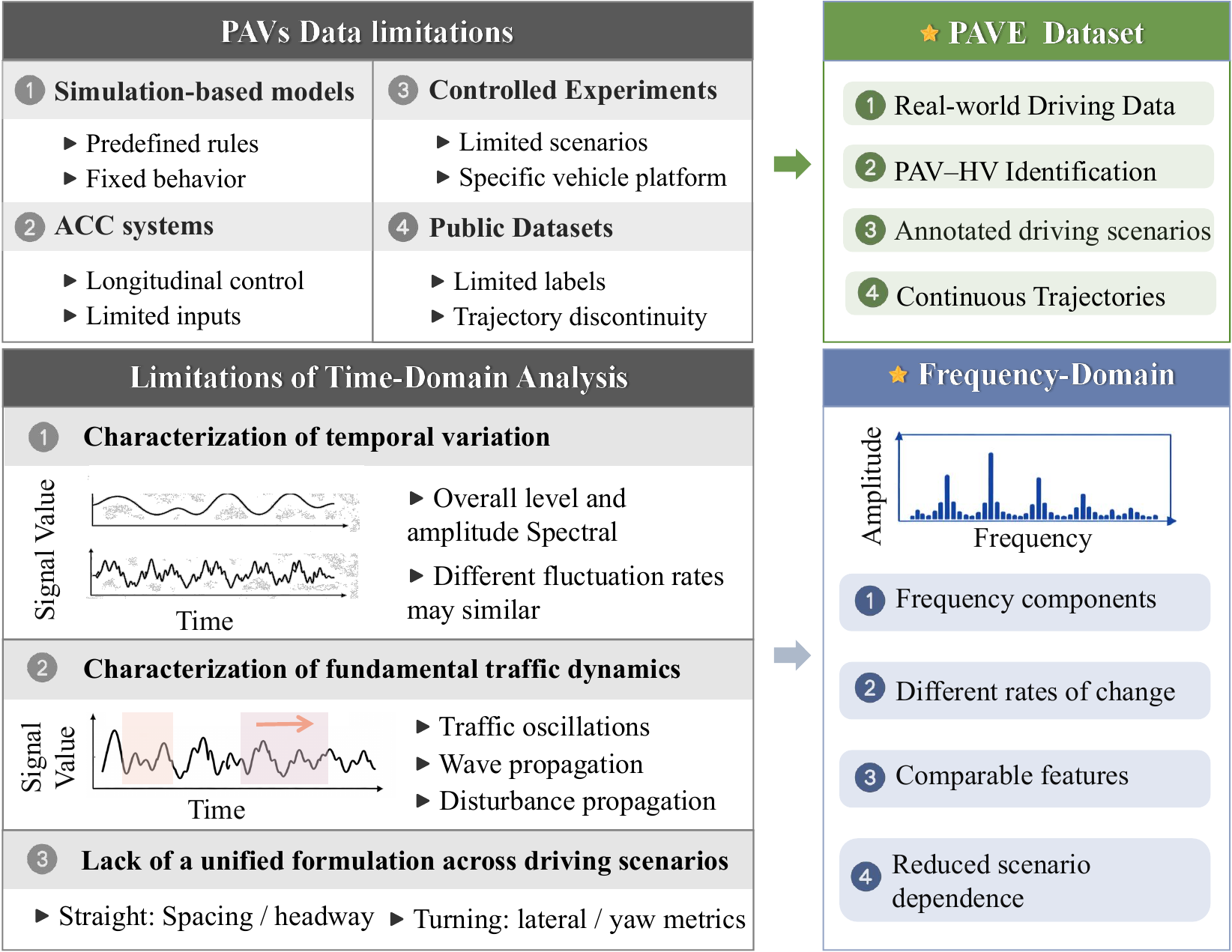}
\caption{Overview of data and methodological gaps in PAV--HV kinematic comparison and the corresponding solutions.}
\label{fig:intro_gap}
\end{figure}

Frequency-domain analysis provides a complementary perspective for characterizing vehicle kinematic characteristics. 
By decomposing time-domain signals into different frequency components, it can distinguish slow and rapid variations in kinematic signals.
A common set of spectral features can then be extracted from different kinematic signals, enabling consistent comparison across signals and reducing reliance on scenario-specific behavioral metrics.
Such approaches have been widely applied to traffic flow oscillations and dynamic traffic states~\cite{Yao2020,Zheng2011Wavelet}, vehicle dynamics and ride comfort ~\cite{Cossalter2006,Griffin2007}, and lateral dynamics and longitudinal disturbance propagation characteristics~\cite{Reddy2001,Gunter2020String,Sun2020}. 
However, frequency-domain analysis has rarely been applied to quantify PAV--HV kinematic differences.

To address these limitations, this study develops a scenario-dependent framework for quantifying PAV--HV kinematic differences using real-world data. 
The framework leverages the PAVE dataset~\cite{li2025pave}, which contains trajectories from both PAVs and HVs with annotated driving scenarios. 
Based on this dataset, Fourier transform is applied to vehicle kinematic signals to extract comparable spectral features. 
Kernel density estimation (KDE) characterizes the spectral-feature distributions, and the Wasserstein distance quantifies the differences between PAVs and HVs.
Comparisons are conducted across driving states, lighting conditions, weather conditions, and vehicle-density categories.
The results reveal pronounced scenario dependence in PAV--HV spectral differences, with the magnitude and direction of the differences varying across driving states, environmental conditions, kinematic variables, and vehicle platforms.

Figure~\ref{fig:intro_gap} summarizes the key data and methodological gaps in existing PAV--HV kinematic comparisons and the corresponding solutions adopted in this study.
The remainder of this paper is organized as follows. 
Section 2 describes the real-world data and the identification of driving scenarios. 
Section 3 presents the signal construction, preprocessing, and frequency-domain feature extraction methods. 
Section 4 presents the results and discussion of the quantified PAV--HV spectral differences across driving scenarios and kinematic variables. 
Section 5 summarizes the conclusions and discusses future research directions.

\section{Dataset and Scenario Extraction}

\label{sec:data}

\subsection{Dataset Overview and Trajectory Reconstruction}

This study uses the PAVE dataset~\cite{li2025pave}, which is derived from an internal GNSS-based vehicle data repository and contains real-world driving records of PAVs and HVs. 
The dataset provides vehicle identifiers, driving-mode labels, vehicle-platform information, scenario annotations, and short trajectory segments collected under diverse driving conditions.
The PAVE data are organized as 11-s key-frame segments designed primarily for short-horizon perception and planning tasks. 
For the frequency-domain analysis in this study, longer continuous trajectories are required, particularly for free-flow cruising and car-following conditions, to provide sufficient temporal support for spectral estimation. 
Therefore, the corresponding vehicle-state records were retrieved from the underlying GNSS database, and continuous trajectories were reconstructed by matching vehicle identifiers and timestamps.

The original GNSS records were collected at 20~Hz and were subsequently processed to a uniform sampling frequency of \(f^s=10\)~Hz, corresponding to a sampling interval of \(\Delta t=0.1\)~s. 
The reconstructed trajectories contain vehicle position, speed, heading angle, vehicle type, platform identity, and scenario information. 
Speed and yaw rate are used as the primary kinematic inputs for subsequent signal construction and frequency-domain analysis, with yaw rate derived from the reconstructed heading angle for turning segments.

Let \(i\) denote the trajectory-segment index.
For trajectory segment \(i\), let \(\mathcal{K}_i:=\{1,\ldots,K_i\}\) denote its sample-index set, where \(K_i\in\mathbb{N}^{+}\) is the number of samples in the segment.
The corresponding time at sample \(k\in\mathcal{K}_i\) is denoted by \(t_{i,k}\).
For all analyzed scenarios except the left-turn and right-turn driving states, \(v_{i,k}\), \(a_{i,k}\), and \(j_{i,k}\) denote the vehicle speed, acceleration, and jerk at time \(t_{i,k}\), respectively.
For turning segments, \(\omega_{i,k}\), \(\alpha_{i,k}\), and \(\eta_{i,k}\) denote the vehicle yaw rate, yaw acceleration, and yaw jerk at time \(t_{i,k}\), respectively.

\subsection{Scenario Definition and Segment Selection}

The trajectory data are characterized using four scenario dimensions provided by the dataset, including driving state, lighting condition, weather condition, and vehicle density.
The corresponding categories and definitions are summarized in Table~\ref{tab:scenario_definition}. 
For turning maneuvers, additional constraints based on heading changes and turning geometry are applied to exclude gradual roadway curves and irregular heading fluctuations. 
The extracted turning segments are retained as complete maneuver episodes, and the sign of the accumulated heading change is used to distinguish left turns from right turns.

\begin{table}[!htb]
\centering
\caption{Scenario dimensions and category definitions based on the PAVE dataset~\cite{li2025pave}.}
\label{tab:scenario_definition}
\small
\scriptsize
\setlength{\tabcolsep}{4pt}
\begin{tabularx}{\columnwidth}{@{}l l X@{}}
\toprule
\textbf{Dimension} & \textbf{Category} & \textbf{Definition} \\
\midrule

\multirow{4}{*}{Driving state}
& Car-following & A preceding vehicle within 25~m and located in the ego vehicle's lane \\
& Free-flow cruising & No preceding vehicle within 25 m in the ego vehicle’s lane \\
& Left turn & Complete left-turn maneuver identified from heading changes \\
& Right turn & Complete right-turn maneuver identified from heading changes \\
\midrule

\multirow{2}{*}{Lighting}
& Day & Sufficient natural illumination, including overcast daytime \\
& Night-lit & Clear artificial illumination during nighttime \\
\midrule

\multirow{2}{*}{Weather}
& Clear & Stable conditions without precipitation \\
& Rain & Active rainfall or post-rain wet-road conditions \\
\midrule

\multirow{3}{*}{Vehicle density}
& Low & 0--5 surrounding vehicles \\
& Medium & 6--15 surrounding vehicles \\
& High & More than 15 surrounding vehicles \\
\bottomrule
\end{tabularx}
\end{table}

\begin{table}[!htb]
\centering
\caption{Distribution of trajectory segments across scenario dimensions. Values are reported as the number of valid trajectory segments, with the corresponding cumulative duration in parentheses (h).}
\label{tab:scenario_distribution}
\scriptsize
\setlength{\tabcolsep}{4pt}
\begin{tabular}{llccccc}
\toprule
Scenario & Category & AV-1 & AV-2 & AV-3 & AV-4 & HV \\
\midrule
Driving state & Cruise & 35 (0.81) & 38 (1.23) & 8 (0.20) & 71 (1.84) & 31 (0.75) \\
& Follow & 51 (1.42) & 149 (5.11) & 128 (3.40) & 163 (4.83) & 104 (2.57) \\
& Left turn & 19 (0.04) & 10 (0.03) & 18 (0.05) & 60 (0.15) & 224 (0.60) \\
& Right turn & 18 (0.04) & 10 (0.03) & 18 (0.04) & 43 (0.10) & 233 (0.63) \\
\midrule
Lighting & Day & 121 (3.61) & 216 (10.46) & 237 (6.66) & 316 (15.20) & 403 (13.72) \\
& Night & 80 (2.32) & 7 (0.25) & 37 (1.10) & 90 (4.46) & 65 (1.94) \\
\midrule
Weather & Clear & 14 (0.37) & 72 (2.74) & 51 (1.35) & 93 (2.90) & 101 (2.86) \\
& Rain & 64 (1.69) & 48 (1.98) & 32 (0.95) & 138 (5.30) & 118 (3.59) \\
\midrule
Vehicle density & Low & 60 (1.52) & 20 (0.58) & 19 (0.55) & 152 (5.44) & 120 (3.52) \\
& Medium & 33 (0.78) & 118 (4.08) & 64 (1.45) & 96 (2.49) & 75 (1.73) \\
& High & 34 (0.99) & 63 (1.78) & 62 (1.49) & 69 (2.04) & 52 (1.32) \\
\bottomrule
\end{tabular}
\end{table}

To provide sufficient samples for subsequent spectral analysis, each continuous segment is required to have a minimum duration of 60 s for all scenarios except left and right turns.
Table~\ref{tab:scenario_distribution} summarizes the number of retained trajectory segments and their cumulative durations for each vehicle group across the four scenario dimensions.
The number of trajectory samples for left-turn and right-turn scenarios is relatively small for some PAV platforms, as drivers in the dataset frequently disengage the automated system and revert to manual control during turning maneuvers.
This tendency may be attributed to drivers' lack of trust in the AV system or to the relatively slow turning behavior of the PAV.
Detailed segment-duration statistics, including the median, first and third quartiles, and interquartile range (IQR), are provided in Appendix~\ref{app:duration_statistics}.
AV trajectories were retained separately for four commercial vehicle platforms (AV--1, AV--2, AV--3, and AV--4), whereas HV trajectories were pooled within each scenario category to construct the reference distributions for PAV--HV comparison.
The four PAV platforms correspond to Toyota bZ3X, Xiaomi Auto YU7, NIO ET5, and AITO M7. 
To avoid directly identifying the individual platforms, the mapping between these vehicle models and AV--1 to AV--4 is anonymized throughout the paper.

\section{Methods}

\begin{figure}[!htb]
\centering
\includegraphics[width=0.98\textwidth]{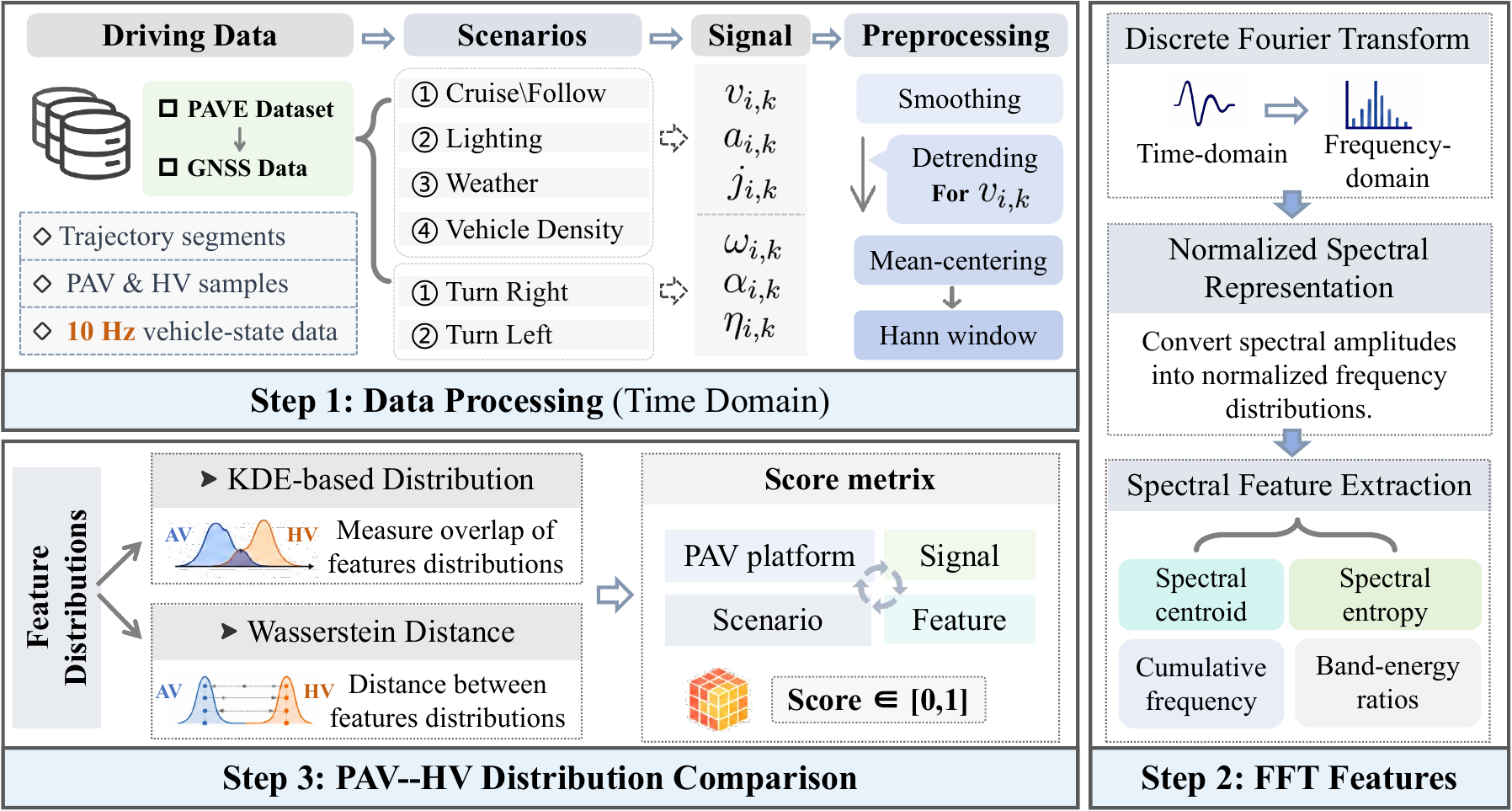}
\caption{Overview of the proposed frequency-domain framework for AVs--HVs kinematic difference quantification..}
\label{fig:workflow}
\end{figure}

To quantify scenario-dependent PAV--HV kinematic differences, this study develops a frequency-domain comparison framework at the trajectory-segment level.
Fig.~\ref{fig:workflow} illustrates the overall workflow.
Processed trajectory segments are first categorized according to four scenario dimensions, including driving state, lighting condition, weather condition, and vehicle density.
For each trajectory segment, scenario-specific kinematic signals are constructed and transformed into the frequency domain after preprocessing.
Several spectral features are extracted, and PAV--HV feature distributions are compared using KDE overlap scores and Wasserstein-based scores across AV platforms, scenarios, signals, and features.

\subsection{Kinematic Signal Construction}
\label{sec:signal}

For trajectory segment \(i\), a kinematic signal is defined as the sampled time series of a kinematic variable over the segment.
The speed signal \(v_{i,k}\) is first smoothed using a centered three-sample moving average to reduce small measurement fluctuations before numerical differentiation.
The smoothed speed at sample \(k\) is denoted by \(\bar{v}_{i,k}\) and calculated according to Equation~\ref{eq:smoothed_speed}.
\begin{equation}
\bar{v}_{ik}=\frac{v_{i,k-1}+v_{i,k}+v_{i,k+1}}{3}.
\label{eq:smoothed_speed}
\end{equation}

Let \(D(\cdot)\) denote the centered finite-difference operator. 
For a generic kinematic signal \(x_{i,k}\), its derivative at sample \(k\) is defined in Equation~\ref{eq:finite_difference}.
\begin{equation}
D(x_{ik})=\frac{x_{i,k+1}-x_{i,k-1}}{2\Delta t}.
\label{eq:finite_difference}
\end{equation}

Based on the smoothed speed signal, \(a_{i,k}\) and \(j_{i ,k}\) are obtained through successive differentiation, where \(a_{i,k}=D(\bar{v}_{i,k})\) and \(j_{i,k}=D(a_{i,k})\). Accordingly, the kinematic signals used for the corresponding scenarios are \(v_{i,k}\), \(a_{i,k}\), and \(j_{i,k}\).
For turning segments, \(\alpha_{i,k}\) and \(\eta_{i,k}\) are obtained from the yaw-rate signal, where \(\alpha_{i,k}=D(\omega_{i,k})\) and \(\eta_{i,k}=D(\alpha_{i,k})\). Accordingly, the kinematic signals used for turning analysis are \(\omega_{i,k}\), \(\alpha_{i,k}\), and \(\eta_{i,k}\).
These constructed kinematic signals are subsequently used as the time-domain inputs for frequency-domain analysis.

\subsection{Frequency-Domain Transformation}

Frequency-domain transformation is performed independently for each kinematic signal of each trajectory segment. 
For all analyzed scenarios except the left-turn and right-turn driving states, the smoothed speed signal is detrended before spectral transformation to remove slow temporal variations that are not the focus of the spectral analysis. 
For trajectory segment \(i\), the fitted polynomial trend of the smoothed speed signal is denoted by \(b_i(t)\) and defined in Equation~\ref{eq:fitted_trend}.
\begin{equation}
b_i(t)=\sum_{r=0}^{m_i}c_{ir}t^r,
\label{eq:fitted_trend}
\end{equation}
\noindent
where \(m_i\) denotes the polynomial degree for segment \(i\), \(r\) denotes the polynomial-order index, and \(c_{i,r}\) denotes the coefficient of the \(r\)-th-order term estimated by least-squares fitting.

The detrended speed \(v^{\mathrm{d}}_{i,k}\) at sample \(k\in\mathcal{K}_i\) is defined as \(v^{\mathrm{d}}_{i,k}=\bar{v}_{i,k}-b_i(t_{i,k})\).
Only the speed signal is detrended. The acceleration and jerk signals constructed from the smoothed speed signal are retained without polynomial detrending. 

For the speed signal, \(x_{i,k}=v^{\mathrm{d}}_{i,k}\), whereas for the other constructed kinematic signals, \(x_{i,k}\) denotes the corresponding signal value without polynomial detrending.
The signal is first mean-centered according to Equation~\ref{eq:mean_center}.
\begin{equation}
\tilde{x}_{i,k}=x_{i,k}-\frac{1}{K_i}\sum_{\ell=1}^{K_i}x_{i\ell},
\label{eq:mean_center}
\end{equation}
\noindent
where \(\ell\) denotes the summation index over the samples of trajectory segment \(i\).

A Hann window is subsequently applied to reduce spectral leakage caused by the finite segment duration, as given in Equation~\ref{eq:hann_window}.
\begin{equation}
x^{\mathrm{win}}_{i,k}=\tilde{x}_{i,k}\frac{1}{2}\left(1-\cos\frac{2\pi(k-1)}{K_i-1}\right).
\label{eq:hann_window}
\end{equation}

The discrete Fourier transform of the windowed signal is then computed using Equation~\ref{eq:dft}.
\begin{equation}
F_{ih}=\sum_{k=1}^{K_i}x^{\mathrm{win}}_{i,k}\exp\left(-\operatorname{j}\frac{2\pi h(k-1)}{K_i}\right),\qquad f_{ih}=\frac{hf^s}{K_i},
\label{eq:dft}
\end{equation}
\noindent
where \(h\) denotes the frequency-bin index, \(F_{ih}\) denotes the Fourier coefficient of segment \(i\) at frequency bin \(h\), \(f_{ih}\) denotes the corresponding frequency, \(f^s=10\)~Hz is the sampling frequency, and \(\operatorname{j}\) denotes the imaginary unit.

\begin{figure}[!htb]
\centering
\includegraphics[width=0.95\linewidth]{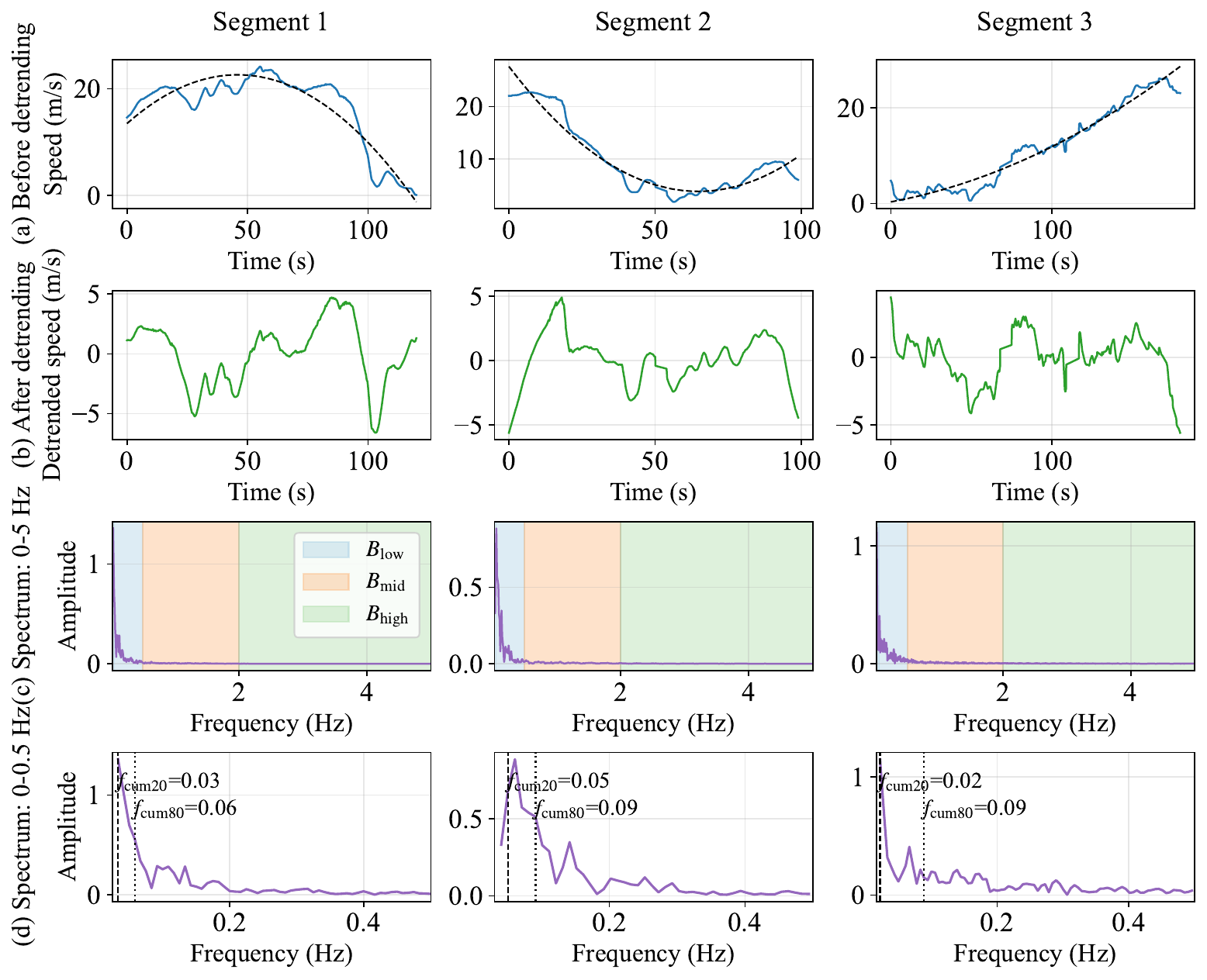}
\caption{Representative examples of speed detrending and spectral representation. 
(a) Smoothed speed signals with fitted trends; (b) detrended speed signals after trend removal; (c) corresponding spectra over 0--5~Hz with different frequency bands; and (d) enlarged spectra over 0--0.5~Hz showing the cumulative-power frequencies \(f_{\mathrm{cum20}}\) and \(f_{\mathrm{cum80}}\).}
\label{fig:detrend_example}
\end{figure}

Only the one-sided spectrum is retained, and the spectral power of segment \(i\) at frequency bin \(h\) is defined as \(P_{ih}=|F_{ih}|^2\).
Fig.~\ref{fig:detrend_example} illustrates the speed preprocessing and frequency-domain transformation, including polynomial detrending and the resulting spectral representation.
The spectral examples also illustrate the frequency bands and cumulative-power frequencies used for subsequent feature extraction, which are formally defined in the next section.

To ensure sufficient temporal support for the retained low-frequency components, let \(T_i\) denote the duration of trajectory segment \(i\). 
Equation~\ref{eq:min_frequency} defines the minimum retained frequency for trajectory segment \(i\).
\begin{equation}
f^{\mathrm{min}}_i=\frac{N_{\mathrm{cyc}}}{T_i},
\label{eq:min_frequency}
\end{equation}
\noindent
where \(N_{\mathrm{cyc}}\) denotes the minimum required number of cycles within a trajectory segment. 
In this study, \(N_{\mathrm{cyc}}=3\) is used for all analyzed scenarios except the left-turn and right-turn driving states, whereas \(N_{\mathrm{cyc}}=1.5\) is used for turning segments.
The upper frequency bound is the Nyquist frequency \(f_s/2=5\)~Hz.

Let \(\mathcal{H}_i\) denote the set of retained frequency-bin indices satisfying \(f^{\mathrm{min}}_i\leq f_{ih}\leq f^s/2\). 
The spectral power is normalized over the retained frequency bins using Equation~\ref{eq:normalized_power}.
\begin{equation}
\rho_{ih}=\frac{P_{ih}}{\sum_{\ell\in\mathcal{H}_i}P_{i\ell}},\qquad h\in\mathcal{H}_i,
\label{eq:normalized_power}
\end{equation}
\noindent
where \(\rho_{ih}\) denotes the normalized spectral power contribution of frequency bin \(h\) for trajectory segment \(i\).

\subsection{Spectral Feature Extraction}
\label{sec:feature}

Based on the normalized spectral power distribution \(\rho_{ih}\), seven spectral features are extracted for each trajectory segment and kinematic signal. 
These features characterize the frequency location, spectral dispersion, cumulative spectral-power distribution, and frequency-band energy allocation.

The spectral centroid \(C_i\) characterizes the overall frequency location of spectral power, providing a measure of whether the spectral energy of a kinematic signal is concentrated toward lower or higher frequencies. Its value is calculated using Equation~\ref{eq:spectral_centroid}.
\begin{equation}
C_i=\sum_{h\in\mathcal{H}_i}f_{ih}\rho_{ih}.
\label{eq:spectral_centroid}
\end{equation}

The normalized spectral entropy \(E_i\) characterizes the dispersion of spectral power across the retained frequency bins, thereby capturing whether the spectral energy of a kinematic signal is concentrated within a limited frequency range or distributed more broadly across frequencies. 
Equation~\ref{eq:spectral_entropy} gives the normalized spectral entropy as
\begin{equation}
E_i=-\frac{1}{\log|\mathcal{H}_i|}\sum_{h\in\mathcal{H}_i}\rho_{ih}\log\rho_{ih}.
\label{eq:spectral_entropy}
\end{equation}

The cumulative frequency features \(f_{i,\mathrm{cum20}}\) and \(f_{i,\mathrm{cum80}}\) denote the frequencies below which 20\% and 80\% of the cumulative spectral power are reached for trajectory segment \(i\), respectively.
The cumulative normalized spectral power is defined in Equation~\ref{eq:cumulative_spectrum}.
\begin{equation}
G_i(f)=\sum_{\substack{h\in\mathcal{H}_i, f_{ih}\leq f}}\rho_{ih}.
\label{eq:cumulative_spectrum}
\end{equation}
\noindent
where \(f_{i,\mathrm{cum20}}=\min\{f:G_i(f)\geq0.2\}\) and \(f_{i,\mathrm{cum80}}=\min\{f:G_i(f)\geq0.8\}\).

The band-energy ratio \(R_{i,B}\) characterizes the proportion of normalized spectral power distributed within frequency band \(B\) for trajectory segment \(i\). 
Its value is computed using Equation~\ref{eq:band_energy_ratio}.
\begin{equation}
R_{i,B}=\sum_{\substack{h\in\mathcal{H}_i, f_{ih}\in B}}\rho_{ih}.
\label{eq:band_energy_ratio}
\end{equation}
\noindent
Three frequency bands are considered, with \(B_{\mathrm{low}}=[0.05,0.5)\,\mathrm{Hz}\), \(B_{\mathrm{mid}}=[0.5,2.0)\,\mathrm{Hz}\), and \(B_{\mathrm{high}}=[2.0,5.0]\,\mathrm{Hz}\).

The extracted frequency-domain features from each trajectory segment and kinematic signal are subsequently used as inputs for the AV--HV spectral difference measures introduced in the next step.

\subsection{PAV--HV Difference Measures}
\label{sec:difference}

PAV--HV spectral differences are quantified at the feature-distribution level because trajectory segments from the two groups are not paired on a one-to-one basis.
For each combination of PAV platform, driving scenario, kinematic signal, and spectral feature, the corresponding PAV and HV feature distributions are compared.
Two complementary distribution-based scores are used, namely the KDE-overlap score \(S^{\mathrm{KDE}}\) and the Wasserstein-based score \(S^{W}\).

For notational simplicity, consider one fixed combination of PAV platform, driving scenario, kinematic signal, and spectral feature.
Let \(Z^{\mathrm{AV}}\) and \(Z^{\mathrm{HV}}\) denote the corresponding PAV and HV feature sample sets extracted from individual trajectory segments.
HV samples are pooled within each scenario--signal--feature combination, whereas PAV samples are evaluated separately for each PAV platform.

The KDE-overlap score \(S^{\mathrm{KDE}}\) characterizes the overlap between the AV and HV feature distributions and therefore reflects their distributional agreement.
Gaussian kernel density estimation is applied separately to \(Z^{\mathrm{AV}}\) and \(Z^{\mathrm{HV}}\), with the bandwidth determined using Scott's rule.
Let \(\hat{p}^{\mathrm{AV}}(z)\) and \(\hat{p}^{\mathrm{HV}}(z)\) denote the corresponding estimated probability density functions.
The KDE-overlap score is computed according to Equation~\ref{eq:kde_overlap}.
\begin{equation}
S^{\mathrm{KDE}}=\int \min\left\{\hat{p}^{\mathrm{AV}}(z),\hat{p}^{\mathrm{HV}}(z)\right\}\mathrm{d}z.
\label{eq:kde_overlap}
\end{equation}
\noindent
The integral is numerically evaluated over the combined feature-value range using trapezoidal integration.
The KDE-overlap score satisfies \(S^{\mathrm{KDE}}\in[0,1]\), with larger values indicating greater overlap between the PAV and HV feature distributions.

The Wasserstein-based score \(S^{W}\) characterizes the separation between the PAV and HV feature distributions.
The first-order Wasserstein distance measures the distributional separation by integrating the absolute differences between the corresponding quantiles of the two distributions.
Let \(Q_{\mathrm{AV}}(u)\) and \(Q_{\mathrm{HV}}(u)\) denote the quantile functions of the empirical PAV and HV feature distributions, respectively, where \(u\in[0,1]\) denotes the cumulative probability level.
The first-order Wasserstein distance \(W_1\)  is computed according to Equation~\ref{eq:wasserstein_raw}.
\begin{equation}
W_1=\int_{0}^{1}\left|Q_{\mathrm{AV}}(u)-Q_{\mathrm{HV}}(u)\right|\mathrm{d}u.
\label{eq:wasserstein_raw}
\end{equation}
\noindent

A larger \(W_1\) indicates greater separation between the two feature distributions, whereas \(W_1=0\) indicates identical distributions.

Because the spectral features have different numerical scales, the Wasserstein distance is normalized by the standard deviation of the corresponding HV feature samples, as shown in Equation~\ref{eq:wasserstein_normalized}.
\begin{equation}
D^{W}=\frac{W_1}{\sigma^{\mathrm{HV}}},
\label{eq:wasserstein_normalized}
\end{equation}
\noindent
where \(\sigma^{\mathrm{HV}}\) denotes the standard deviation of \(Z^{\mathrm{HV}}\).
This normalization expresses the PAV--HV distributional separation relative to the variability of the corresponding HV feature distribution.
Feature combinations with zero or undefined \(\sigma^{\mathrm{HV}}\) are excluded from the calculation.

To obtain a bounded similarity score with the same interpretation direction as \(S^{\mathrm{KDE}}\), the normalized Wasserstein distance is computed according to Equation~\ref{eq:wasserstein_score}.
\begin{equation}
S^{W}=\exp\left(-D^{W}\right).
\label{eq:wasserstein_score}
\end{equation}
\noindent
Accordingly, \(S^{W}\in(0,1]\), where values closer to 1 indicate greater similarity between the PAV and HV feature distributions.

Both \(S^{\mathrm{KDE}}\) and \(S^{W}\) are calculated separately for each PAV platform, driving scenario, kinematic signal, and spectral feature.

\subsection{Robustness Assessment}

The quantified PAV--HV spectral differences may be affected by sampling variability due to the finite number of trajectory segments.
To evaluate the statistical uncertainty of the proposed spectral comparison framework, bootstrap confidence intervals were calculated for the resulting comparison scores.

Bootstrap confidence intervals were estimated by resampling the pre-computed frequency-domain feature samples with replacement within each scenario--signal--vehicle combination. 
For each bootstrap resample, the PAV--HV comparison scores were recalculated following the same procedure described above. 
A total of 1000 bootstrap resamples were generated for each combination, and the 95\% confidence intervals were obtained using the percentile method from the resulting empirical bootstrap distributions.

\section{Results}

\subsection{Overview of Scenario-Dependent PAV--HV Kinematic Differences}
The PAV--HV spectral differences are evaluated separately for each PAV platform to preserve platform-specific behavioral characteristics that may be obscured by pooled analysis. Table~\ref{tab:kde_overall_scores} and Table~\ref{tab:wasserstein_overall_scores} provide an overall summary of the mean KDE overlap and Wasserstein scores across all analyzed scenario branches and PAV platforms. 

\begin{table}[!htb]
\centering
\caption{Mean \(S^{\mathrm{KDE}}\) across the four PAV platforms for different scenario branches. Bold values indicate the highest score within each scenario comparison for each PAV platform.}
\label{tab:kde_overall_scores}
\begin{tabular}{llcccc}
\toprule
Scenario & Branch & AV-1 & AV-2 & AV-3 & AV-4 \\
\midrule
\multirow{4}{*}{Driving state}
& Cruise     & 0.571 & 0.652 & 0.663 & 0.524 \\
& Follow     & \textbf{0.694} & \textbf{0.691} & \textbf{0.707} & \textbf{0.633} \\
\cmidrule(lr){2-6}
& Left Turn  & 0.549 & 0.678 & 0.676 & 0.703 \\
& Right Turn & \textbf{0.682} & \textbf{0.733} & \textbf{0.710} & \textbf{0.735} \\
\midrule
\multirow{2}{*}{Lighting}
& Day   & \textbf{0.697} & \textbf{0.729} & \textbf{0.778} & 0.685 \\
& Night & 0.652 & 0.626 & 0.690 & \textbf{0.733} \\
\midrule
\multirow{2}{*}{Weather}
& Clear & \textbf{0.716} & \textbf{0.766} & \textbf{0.759} & \textbf{0.728} \\
& Rain  & 0.636 & 0.705 & 0.747 & 0.632 \\
\midrule
\multirow{3}{*}{Vehicle density}
& Low  & 0.597 & 0.546 & \textbf{0.697} & 0.619 \\
& Mid  & 0.594 & 0.658 & 0.688 & 0.658 \\
& High & \textbf{0.675} & \textbf{0.661} & 0.683 & \textbf{0.695} \\
\bottomrule
\end{tabular}
\end{table}

\begin{table}[!htb]
\centering
\caption{Mean $S^{W}$ across the four PAV platforms for different scenario branches. Bold values indicate the highest score within each scenario comparison for each PAV platform.}
\label{tab:wasserstein_overall_scores}
\begin{tabular}{llcccc}
\toprule
Scenario & Branch & AV-1 & AV-2 & AV-3 & AV-4 \\
\midrule
\multirow{4}{*}{Driving state}
& Cruise     & 0.458 & \textbf{0.609} & \textbf{0.648} & 0.355 \\
& Follow     & \textbf{0.638} & 0.566 & 0.638 & \textbf{0.461} \\
\cmidrule(lr){2-6}
& Left Turn  & 0.429 & 0.572 & 0.549 & 0.559 \\
& Right Turn & \textbf{0.532} & \textbf{0.605} & \textbf{0.596} & \textbf{0.623} \\
\midrule
\multirow{2}{*}{Lighting}
& Day   & \textbf{0.613} & \textbf{0.632} & \textbf{0.736} & 0.491 \\
& Night & 0.564 & 0.569 & 0.675 & \textbf{0.710} \\
\midrule
\multirow{2}{*}{Weather}
& Clear & \textbf{0.691} & \textbf{0.726} & \textbf{0.771} & \textbf{0.674} \\
& Rain  & 0.542 & 0.662 & 0.730 & 0.436 \\
\midrule
\multirow{3}{*}{Vehicle density}
& Low  & 0.452 & 0.435 & \textbf{0.669} & 0.409 \\
& Mid  & 0.473 & 0.513 & 0.607 & 0.487 \\
& High & \textbf{0.628} & \textbf{0.614} & 0.625 & \textbf{0.613} \\
\bottomrule
\end{tabular}
\end{table}

Detailed results for the driving-state dimension, including cruise--follow and left--right turn comparisons, are reported for all four platforms. For lighting, weather, and vehicle density, the main text focuses on AV-4, which contains the largest amount of trajectory data among the four platforms. Complete platform-specific results across all scenario dimensions, kinematic signals, and frequency-domain features are provided in Appendix~\ref{app:complete_scores}.

\subsection{Cruise and Car-Following Conditions}

This section examines how the PAV--HV spectral differences vary between free-flow cruising and car-following conditions.
A clear platform-dependent pattern is observed across the two comparison metrics.
For AV-1 and AV-4, car-following generally yields higher \(S^{\mathrm{KDE}}\) and \(S^{W}\) scores for speed and acceleration, indicating smaller PAV--HV spectral differences than under free-flow cruising.
In contrast, AV-2 and AV-3 show substantially weaker and less consistent cruise--follow changes, with the direction of the differences depending on the kinematic signal and comparison metric.
Across platforms, the cruise--follow contrast is most pronounced for speed, remains evident but less consistent for acceleration, and is comparatively limited for jerk.

\begin{figure}[!htb]
\centering
\includegraphics[width=\textwidth]{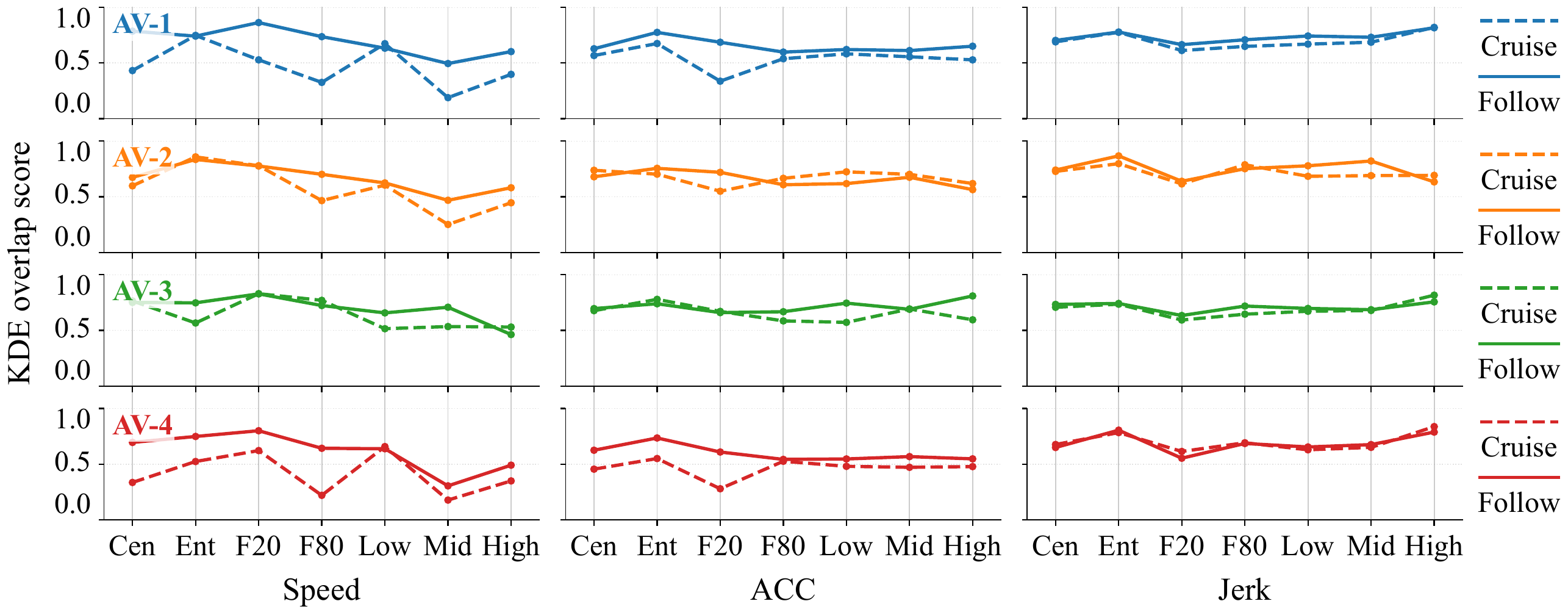}
\caption{Feature-level \(S^{\mathrm{KDE}}\) values for cruise and car-following conditions across the four PAV platforms. Columns represent speed, acceleration (ACC), and jerk, and rows correspond to AV-1--AV-4. The horizontal axis includes spectral centroid (Cen), spectral entropy (Ent), cumulative frequencies F20 and F80, and relative energy in the low-, mid-, and high-frequency bands (Low, Mid, and High). The vertical axis denotes the KDE overlap score \(S^{\mathrm{KDE}}\). Colors identify PAV platforms, while dashed and solid lines indicate cruise and car-following conditions, respectively.}
\label{fig:cf_kde}
\end{figure}

\begin{figure}[!htb]
\centering
\includegraphics[width=\textwidth]{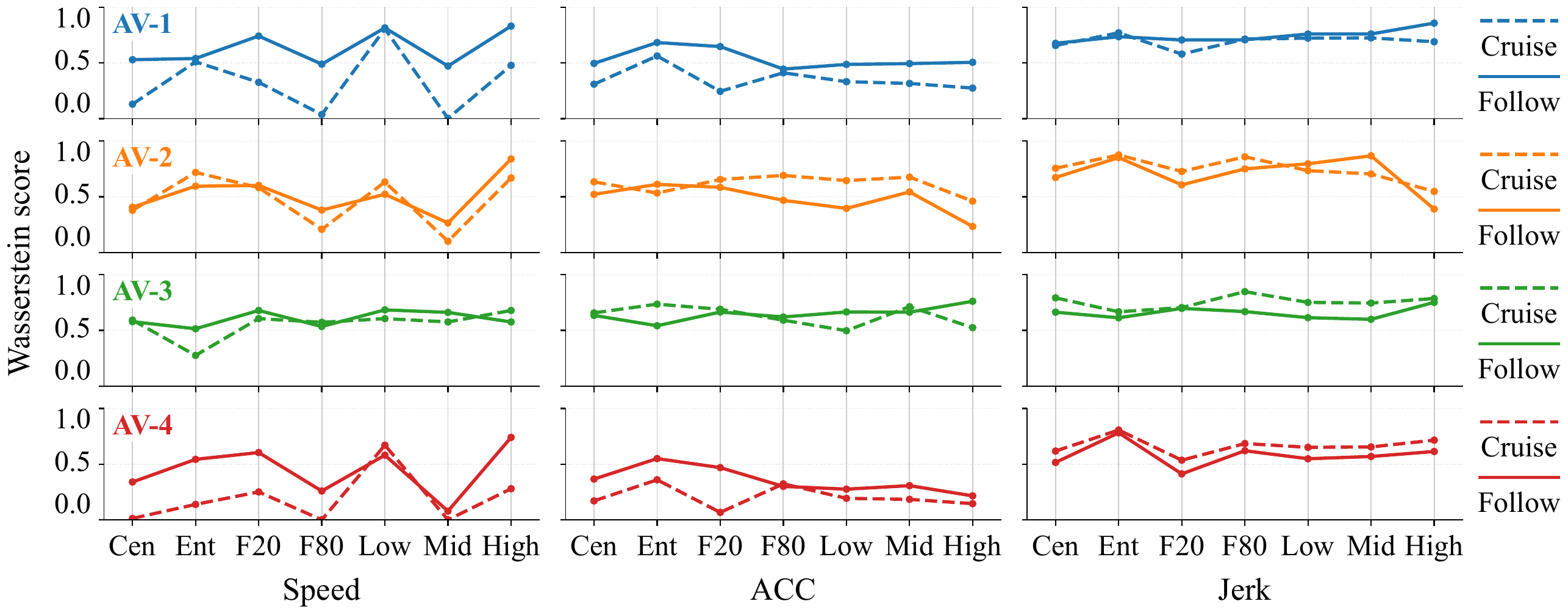}
\caption{Feature-level $S^{W}$ values for cruise and car-following conditions across the four PAV platforms.}
\label{fig:cf_w}
\end{figure}

The feature-level results in Figs.~\ref{fig:cf_kde} and \ref{fig:cf_w} demonstrate clear variations among PAV platforms. 
For several speed-related and acceleration-related features, car-following conditions generally lead to higher PAV--HV agreement compared with free-flow cruising, particularly for AV--1 and AV--4. 
However, this tendency is less pronounced for AV--2 and AV--3, where the changes between cruise and car-following are smaller or vary across frequency-domain features. 
These results indicate that the cruise--follow differences are not consistent across AV platforms and frequency-domain features.

In addition, the two distribution-based measures show consistent qualitative trends but different magnitudes of variation. 
The $S^{\mathrm{KDE}}$ results generally indicate improved PAV--HV distributional agreement under car-following for several speed and acceleration features, whereas the corresponding changes in $S^{W}$ are more pronounced for some features. 
The different responses of the two measures suggest that PAV--HV spectral differences involve multiple aspects of feature distributions and cannot be fully characterized by a single distributional descriptor.

\begin{figure}[!htb]
\centering
\includegraphics[width=\textwidth]{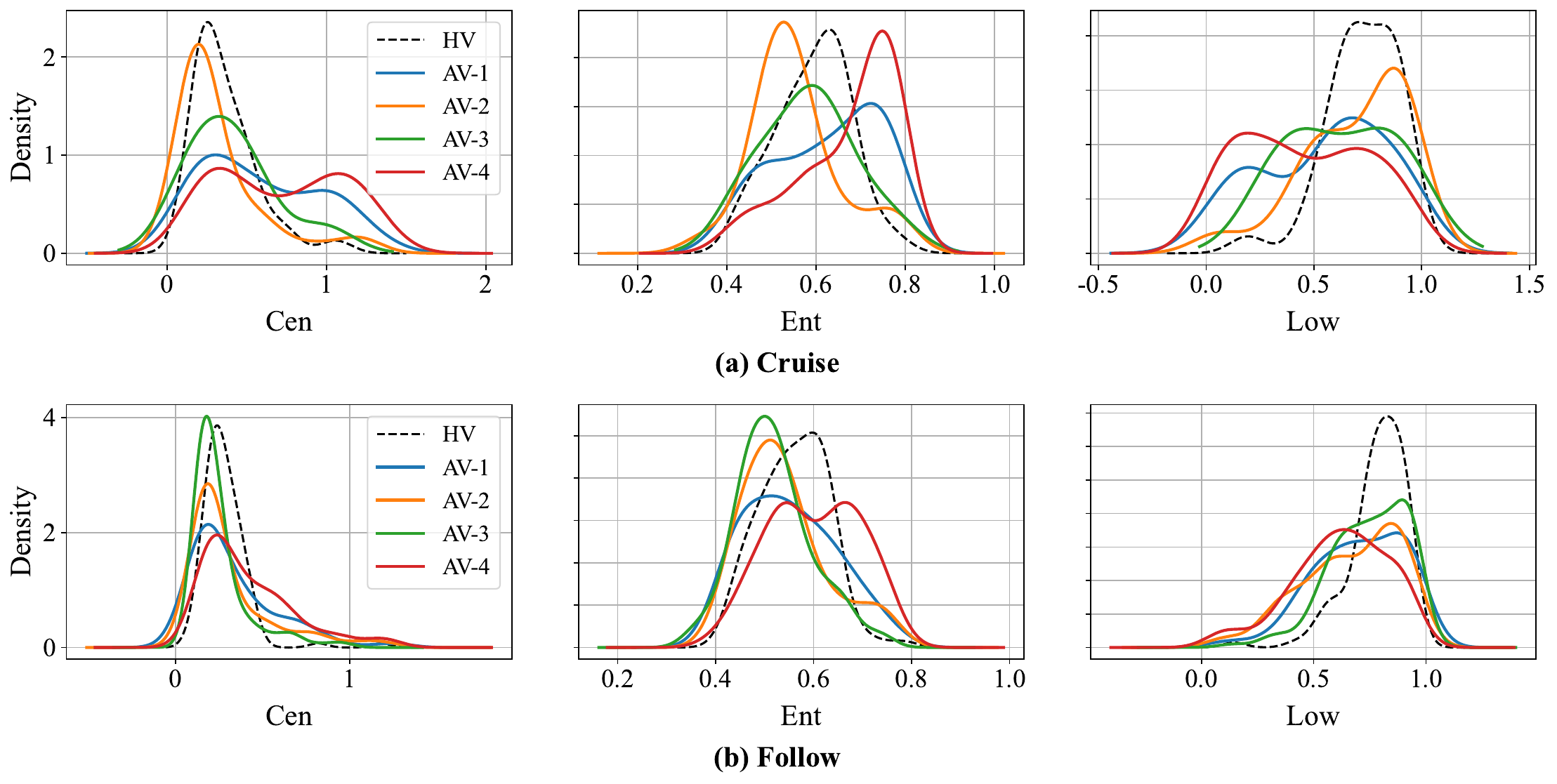}
\caption{KDE distributions of acceleration-related frequency-domain features for four AV platforms under cruise and car-following conditions. }
\label{fig:acc}
\end{figure}

Figure~\ref{fig:acc} presents representative KDE distributions of three acceleration-related spectral features under free-flow cruising and car-following conditions.
The horizontal axis represents the corresponding spectral feature value, while the vertical axis denotes the estimated probability density.
The dashed black curves represent the HV distributions, and the colored solid curves represent the distributions of AV-1 through AV-4.
The upper and lower rows correspond to free-flow cruising and car-following conditions, respectively.
Complete KDE distributions for all kinematic signals and spectral features are provided in Appendix~\ref{app:complete_kde}.
The figure provides a distribution-level visualization of the overlap and separation between PAV and HV spectral features, complementing the quantitative comparison based on \(S^{\mathrm{KDE}}\).

\begin{table}[!htb]
\centering
\footnotesize
\caption{Signal-level PAV--HV spectral scores for AV--4 with 95\% bootstrap confidence intervals. Values in brackets indicate the 95\% confidence intervals obtained from 1,000 trajectory-segment bootstrap resamples.}
\label{tab:av4_cf_robustness}

\textbf{(a) $S^{\mathrm{KDE}}$ for cruise and follow}

\begin{tabular}{lccc}
\toprule
Scenario & Speed & Acceleration & Jerk\\
\midrule
Cruise & 0.412 [0.307--0.475] & 0.463 [0.337--0.572] & 0.697 [0.561--0.751]\\
Follow & 0.617 [0.553--0.679] & 0.596 [0.517--0.662] & 0.687 [0.608--0.740]\\
\bottomrule
\end{tabular}

\vspace{2mm}

\textbf{(b) $S^{W}$ for cruise and follow}

\begin{tabular}{lccc}
\toprule
Scenario & Speed & Acceleration & Jerk\\
\midrule
Cruise & 0.193 [0.111--0.275] & 0.206 [0.073--0.390] & 0.666 [0.475--0.787]\\
Follow & 0.449 [0.350--0.557] & 0.354 [0.176--0.521] & 0.580 [0.451--0.696]\\
\bottomrule
\end{tabular}

\end{table}

The bootstrap analysis further evaluates the robustness of the identified cruise--follow differences using AV--4 as an example. 
As summarized in Table~\ref{tab:av4_cf_robustness}, the increase in speed and acceleration scores under car-following is consistently observed for both $S^{\mathrm{KDE}}$ and $S^{W}$, whereas jerk remains relatively unchanged. 
The bootstrap confidence intervals indicate that the observed trends are generally stable under repeated trajectory-segment resampling, although the uncertainty varies among signals and measures. 
Overall, the robustness analysis confirms that the dominant cruise--follow difference is reflected in speed-related spectral characteristics, while higher-order jerk characteristics exhibit weaker dependence on vehicle interaction conditions.ng interaction.

\subsection{Left-turn and Right-turn Maneuvers}

This section examines how PAV--HV spectral differences vary between left-turn and right-turn maneuvers.
The influence of turning direction differs across signals and PAV platforms, and the differences are more pronounced for yaw acceleration than for yaw rate.
Yaw-rate scores remain broadly comparable between the two turning directions across most PAV platforms, whereas yaw acceleration generally shows higher scores during right turns, particularly for AV-3 and AV-4 under both \(S^{\mathrm{KDE}}\) and \(S^{W}\).
Yaw jerk also exhibits directional differences, although the changes are less consistent across platforms and spectral features.

\begin{figure}[!htb]
\centering
\includegraphics[width=\textwidth]{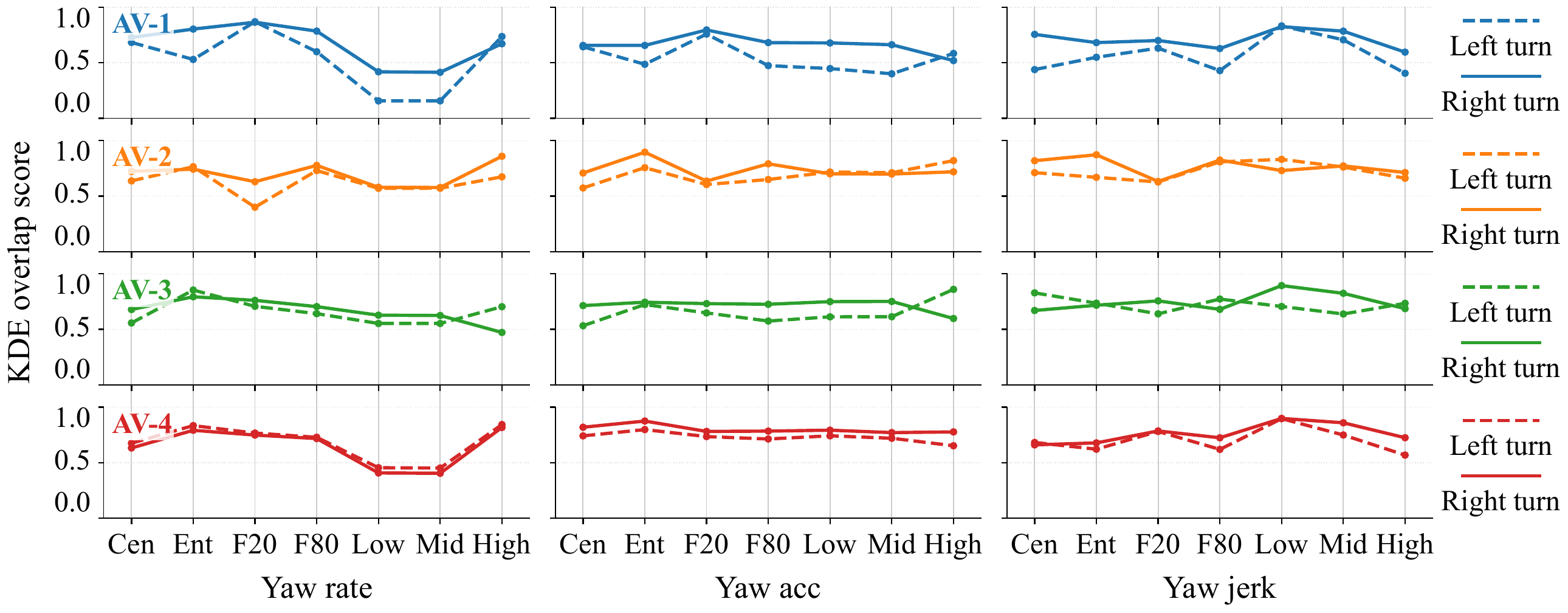}
\caption{Feature-level $S^{\mathrm{KDE}}$ values for left-turn and right-turn conditions across the four PAV platforms.}
\label{fig:turn_kde}
\end{figure}

\begin{figure}[!htb]
\centering
\includegraphics[width=\textwidth]{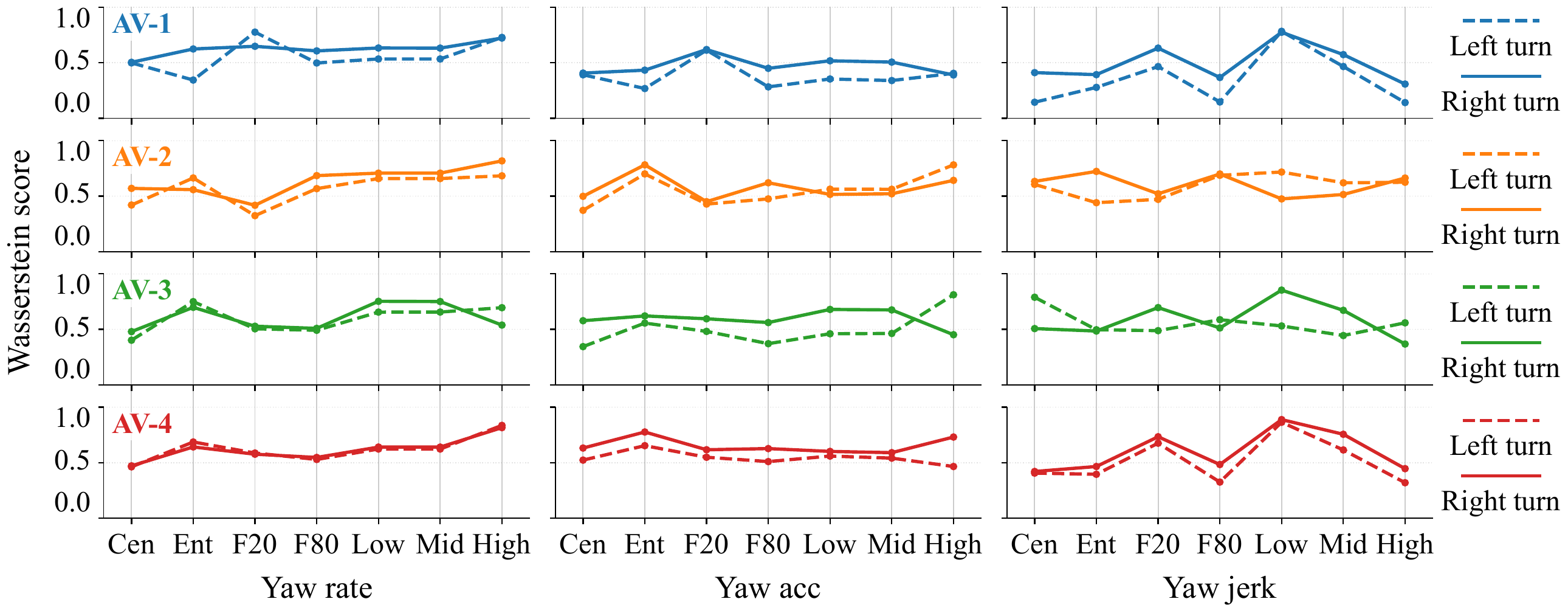}
\caption{Feature-level $S^{W}$ values for left-turn and right-turn conditions across the four PAV platforms.}
\label{fig:turn_wasserstein}
\end{figure}

Figs.~\ref{fig:turn_kde} and ~\ref{fig:turn_wasserstein} present the feature-level \(S^{\mathrm{KDE}}\) and \(S^{W}\) results under left-turn and right-turn conditions. 
The results indicate that the PAV--HV spectral differences are relatively consistent between left and right turns for yaw rate, where both measures exhibit similar score levels across most frequency-domain features. 
In contrast, yaw acceleration shows more noticeable directional variations, with right turns generally achieving higher scores than left turns for multiple features. 
A similar but weaker tendency is observed for yaw jerk, where several frequency-domain features also show higher scores under right-turn conditions. 
These results suggest that turning direction has a greater influence on higher-order yaw dynamics than on the overall yaw rate, leading to more evident directional differences in PAV--HV spectral characteristics.

\begin{figure}[!htb]
\centering
\includegraphics[width=\textwidth]{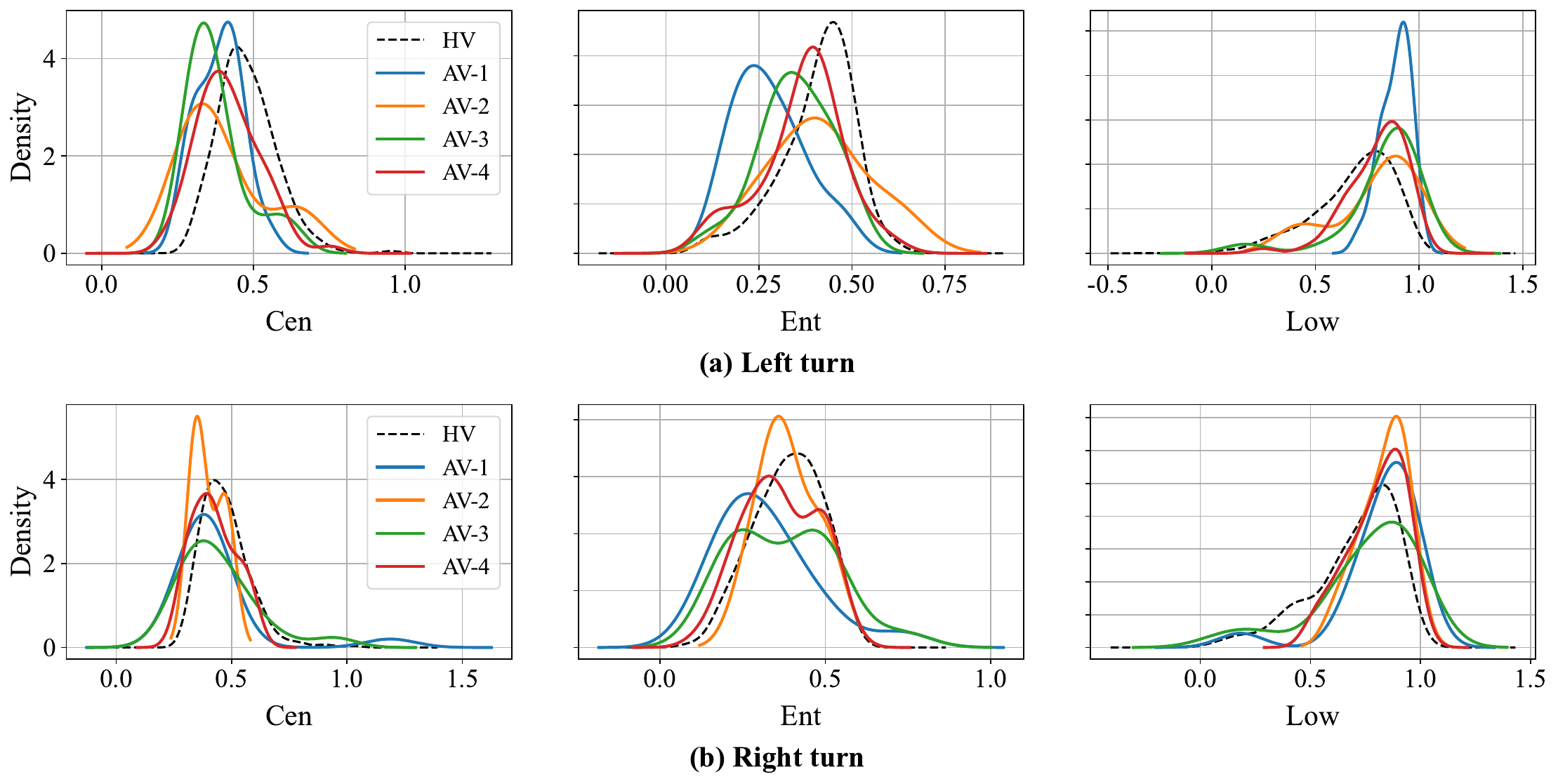}
\caption{KDE distributions of three representative yaw-related spectral features (Cen, Ent, and Low) under left-turn and right-turn maneuvers.}
\label{fig:turn_acc}
\end{figure}

Fig.~\ref{fig:turn_acc} presents representative KDE distributions of three yaw-related spectral features under left-turn and right-turn conditions for HV and the four PAV platforms. Complete KDE distributions for all yaw-related signals, spectral features, and PAV platforms are provided in Appendix~\ref{app:complete_kde}.

\begin{table}[!htb]
\centering
\footnotesize
\caption{Signal-level PAV--HV spectral scores for AV--4 during left- and right-turn maneuvers with 95\% bootstrap confidence intervals. Values in brackets indicate the 95\% confidence intervals obtained from 1,000 trajectory-segment bootstrap resamples.}
\label{tab:av4_turn_robustness}

\textbf{(a) $S^{\mathrm{KDE}}$ for left and right turns}

\begin{tabular}{lccc}
\toprule
Turn & Yaw rate & Yaw acceleration & Yaw jerk\\
\midrule
Left turn & 0.678 [0.592--0.736] & 0.728 [0.647--0.779] & 0.702 [0.633--0.743]\\
Right turn & 0.645 [0.564--0.698] & 0.799 [0.698--0.826] & 0.761 [0.680--0.809]\\
\bottomrule
\end{tabular}

\vspace{2mm}

\textbf{(b) $S^{W}$ for left and right turns}

\begin{tabular}{lccc}
\toprule
Turn & Yaw rate & Yaw acceleration & Yaw jerk\\
\midrule
Left turn & 0.620 [0.546--0.684] & 0.542 [0.454--0.632] & 0.514 [0.428--0.584]\\
Right turn & 0.618 [0.550--0.675] & 0.653 [0.545--0.736] & 0.598 [0.475--0.688]\\
\bottomrule
\end{tabular}

\end{table}

Table~\ref{tab:av4_turn_robustness} further evaluates the stability of the observed turning-direction differences using bootstrap resampling. 
For both \(S^{\mathrm{KDE}}\) and \(S^{W}\), yaw rate shows comparable scores between left and right turns, whereas yaw acceleration and yaw jerk exhibit higher scores under right-turn conditions. 
The bootstrap confidence intervals remain relatively narrow for most signals, indicating that the observed directional variations are stable under repeated trajectory-segment resampling.

\subsection{Lighting and Weather Conditions}

\begin{figure}[!htbp]
\centering
\includegraphics[width=\textwidth]{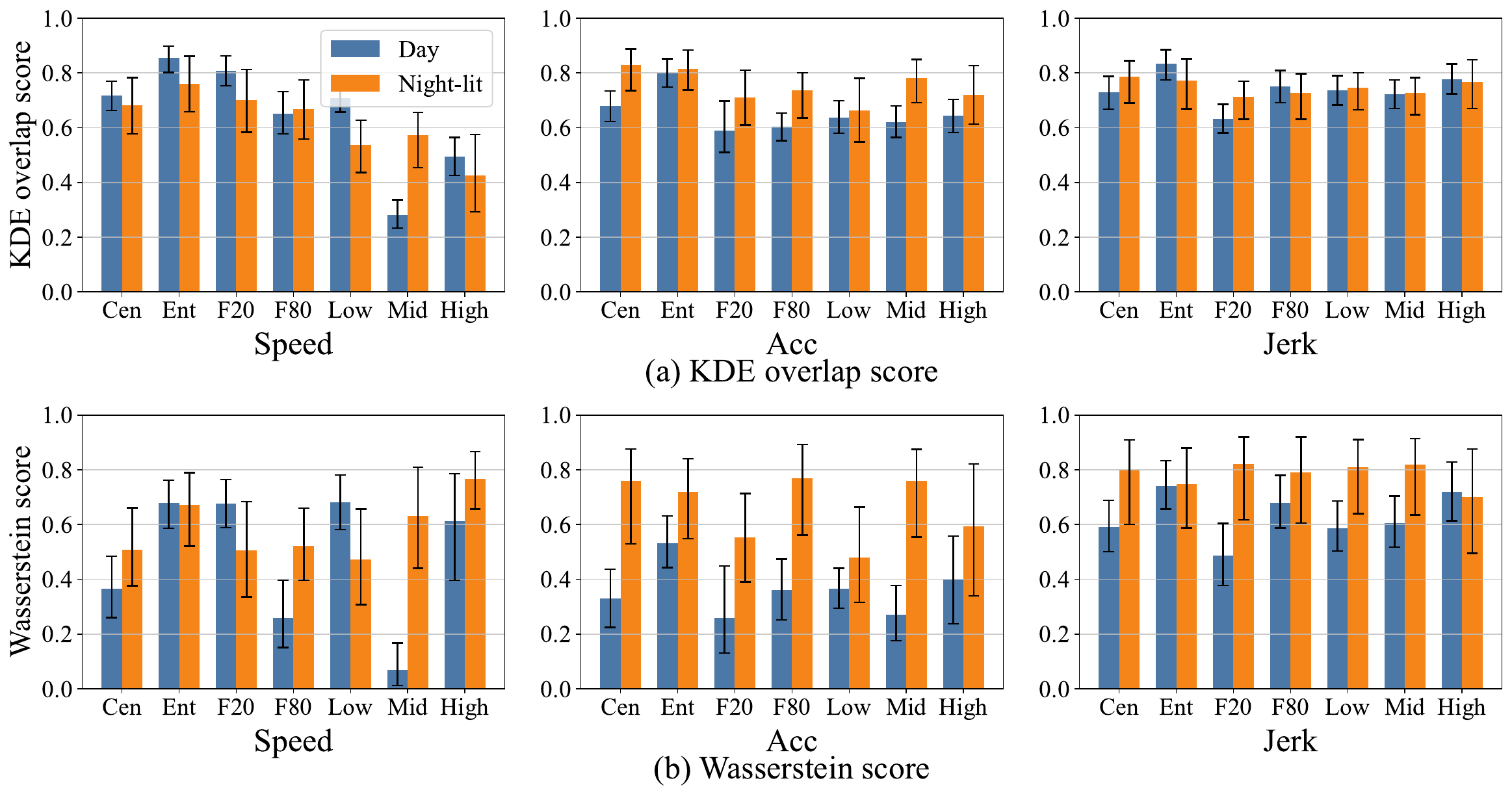}
\caption{Signal-level PAV--HV spectral scores for AV--4 under different lighting conditions. 
The upper and lower panels show $S^{\mathrm{KDE}}$ and $S^{W}$, respectively, for day and night-lit conditions across speed, acceleration, and jerk signals. }
\label{fig:day}
\end{figure}

\begin{figure}[!htbp]
\centering
\includegraphics[width=\textwidth]{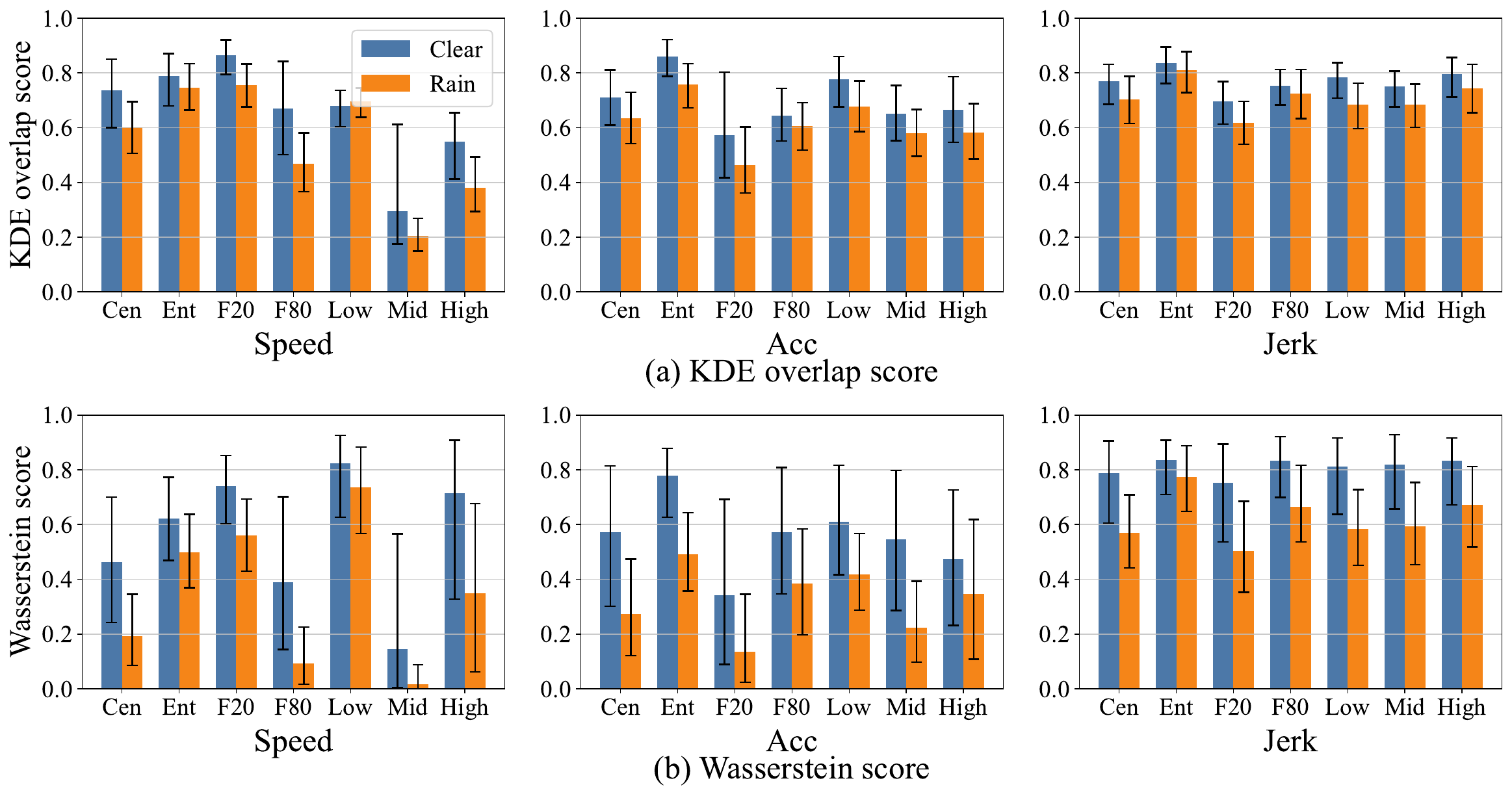}
\caption{Signal-level PAV--HV spectral scores for AV--4 under different weather conditions. 
The upper and lower panels show $S^{\mathrm{KDE}}$ and $S^{W}$, respectively, for clear and rain conditions across speed, acceleration, and jerk signals. }
\label{fig:weather}
\end{figure}

This section examines the influence of environmental conditions, including lighting and weather, on PAV--HV spectral differences. 
Compared with lighting conditions, which produce feature-dependent variations without a consistent directional trend, weather conditions exhibit a clearer influence, with rainy conditions generally associated with lower PAV--HV spectral agreement. 
Figs.~\ref{fig:day} and~\ref{fig:weather} present the signal-level $S^{\mathrm{KDE}}$ and $S^{W}$ results for AV--4 under different lighting and weather conditions, respectively. 
In these figures, the x-axis represents the frequency-domain features, the y-axis indicates the corresponding spectral scores, and the bars compare different environmental conditions with error bars showing the 95\% bootstrap confidence intervals.

For lighting conditions, the difference between day and night-lit conditions is more evident in the $S^{W}$ results than in the $S^{\mathrm{KDE}}$ results. 
The $S^{\mathrm{KDE}}$ scores remain relatively close between the two lighting conditions, although several frequency-domain features exhibit condition-dependent variations. 
In contrast, $S^{W}$ shows larger variations between day and night-lit conditions, particularly for acceleration and jerk signals, where several frequency-domain features achieve higher scores under night-lit conditions.
The speed signal presents more feature-dependent changes, with the relative ranking between the two lighting conditions varying across frequency-domain features. 
These results indicate that the Wasserstein-based score is more sensitive to lighting-related changes in PAV--HV feature distributions than the KDE overlap score.

Weather conditions show a noticeable influence on PAV--HV spectral differences, with rainy conditions generally producing lower $S^{\mathrm{KDE}}$ and $S^{W}$ values than clear conditions. 
For the speed signal, both clear and rainy conditions exhibit relatively low scores in the mid- and high-frequency features, indicating persistent PAV--HV discrepancies in these frequency ranges regardless of weather conditions. 
The larger variations observed in $S^{W}$, particularly for speed and acceleration, suggest that rainy conditions mainly affect the overall displacement of PAV--HV feature distributions rather than only their distribution overlap.

\subsection{Vehicle-Density Conditions}

\begin{figure}[!htbp]
\centering
\includegraphics[width=\textwidth]{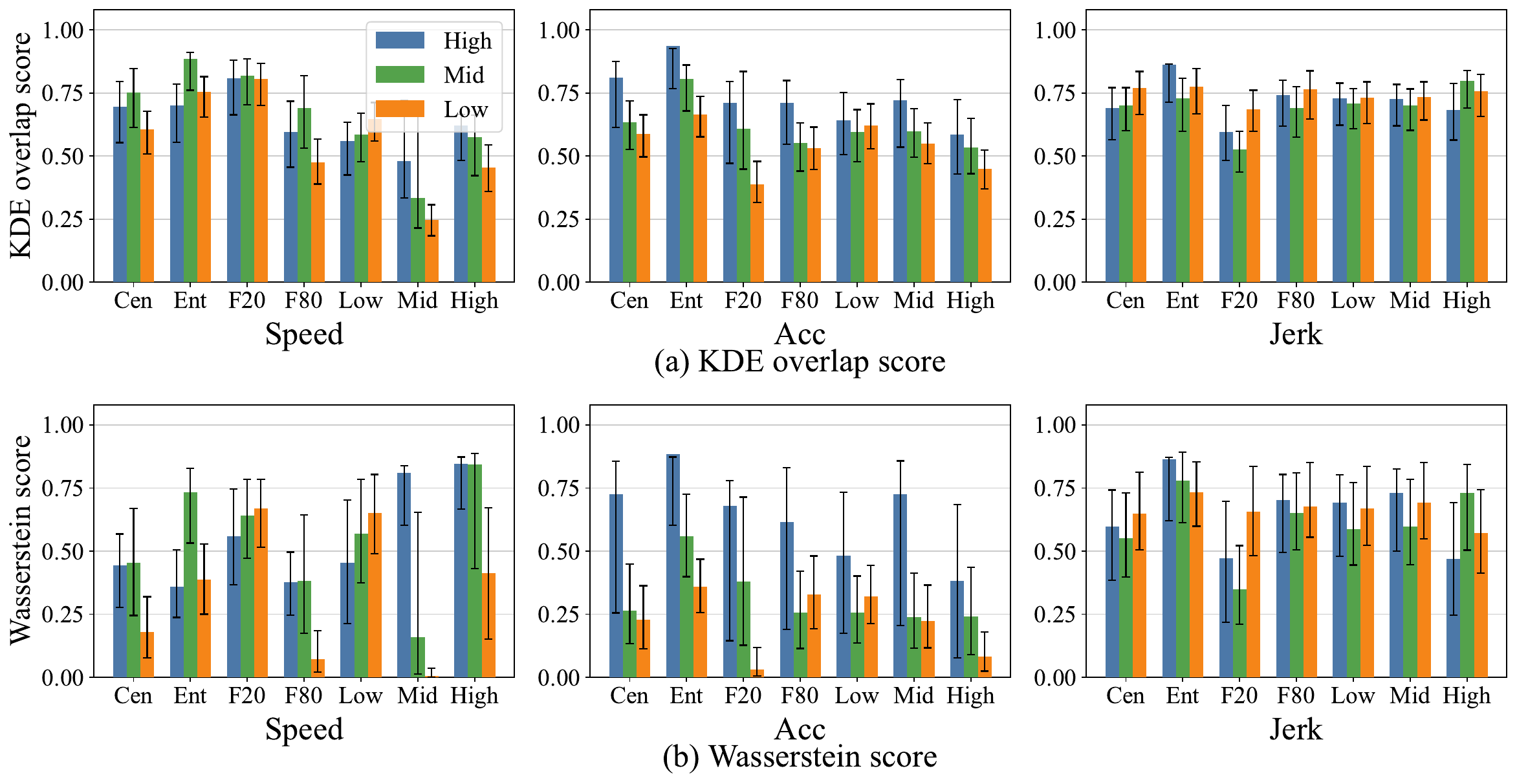}
\caption{Signal-level PAV--HV spectral scores for AV--4 under different vehicle density. 
The upper and lower panels show $S^{\mathrm{KDE}}$ and $S^{W}$, respectively. }
\label{fig:density}
\end{figure}

This section investigates the influence of vehicle density on PAV--HV spectral differences. 
Fig.~\ref{fig:density} presents the signal-level $S^{\mathrm{KDE}}$ and $S^{W}$ results for AV--4 under low, medium, and high vehicle-density conditions. 
The vertical axis indicates the PAV--HV spectral scores, and different colors correspond to different density levels.

The results show that the influence of vehicle density is mainly observed in specific signals and frequency-domain features. 
The acceleration signal demonstrates a clearer response to vehicle-density variations, where both $S^{\mathrm{KDE}}$ and $S^{W}$ show decreasing scores with increasing density for several features, indicating larger PAV--HV spectral differences under higher-density conditions. 
The speed signal exhibits a similar tendency for low- and mid-frequency features, whereas other speed-related features and the jerk signal remain relatively stable across density levels.

The observed density-related variations may be associated with the different vehicle response strategies under constrained traffic conditions. 
As traffic density increases, interactions with surrounding vehicles impose more frequent and less predictable adjustments in vehicle longitudinal behavior. 
The differences in response timing, acceleration control, and adjustment magnitude between PAVs and HVs are therefore more likely to accumulate in speed and acceleration signals, particularly in low- and mid-frequency components. 
In contrast, jerk reflects rapid changes in acceleration, which are influenced by short-duration fluctuations and measurement variability, resulting in weaker and less consistent density-related differences.

The bootstrap confidence intervals are relatively wide for several density-related comparisons, particularly for the Wasserstein-based scores, indicating increased uncertainty in estimating distributional displacement under different density conditions. 
Therefore, the observed density-related variations should be interpreted as signal- and feature-specific differences rather than a uniform effect of vehicle density on PAV--HV spectral characteristics.

\section{Conclusion}

This paper develops a scenario-dependent framework based on real-world data to systematically quantify the kinematic differences between PAVs and HVs across different driving scenarios. The framework extracts comparable spectral features from vehicle kinematic signals through Fourier transform. The distributional differences in spectral features between PAVs and HVs are then quantified with KDE and the Wasserstein distance. The framework is implemented and evaluated on the PAVE dataset, which provides PAVs and HVs trajectories together with driving-scenario annotations for constructing comparative data under different driving conditions. The analysis covers multiple scenario dimensions, including driving states, lighting, weather, and vehicle density, to systematically examine how PAV--HV kinematic differences vary across driving scenarios and kinematic variables.

The quantitative results demonstrate that the proposed frequency-domain framework can provide stable quantification of PAV--HV kinematic differences. The KDE and Wasserstein measures exhibit broadly consistent trends across the major driving scenarios, indicating that the identified differences are not driven by a single distributional measure. Bootstrap confidence intervals further support the statistical reliability of the results.

The results reveal pronounced scenario dependence in PAV--HV spectral differences, with both the magnitude and direction of the differences varying across driving conditions, kinematic variables, and vehicle platforms. Several robust patterns nevertheless emerged across PAVs platforms. Speed showed the most consistent cruise--follow distinction, yaw acceleration and yaw jerk exhibited the clearest left--right turning differences, and acceleration showed the most consistent response to weather conditions. In contrast, the effects of lighting and vehicle density were more platform-dependent, although acceleration generally remained more sensitive to these external conditions than jerk. Overall, the findings indicate that PAV--HV kinematic differences are not governed by a single uniform pattern but are shaped jointly by driving scenario, kinematic variable, and vehicle platform.

This study demonstrates the proposed frequency-domain framework provides a unified approach for characterizing PAVs kinematic behavior and quantifying PAV--HV differences across driving scenarios and kinematic variables. One limitation of this study is that the available data cover only a limited number of PAVs platforms and driving scenarios, with relatively small sample sizes in some conditions. Future work should expand the range of vehicle platforms, road environments, and driving conditions to further examine the generalizability of the identified differences. Another limitation is that the present study focuses on the quantitative characterization of PAV--HV kinematic differences and does not further explain how these differences relate to the decision-making strategies of ADS. Future research could incorporate perception, decision-making, and control information to investigate the mechanisms underlying PAVs kinematic differences and to examine their implications for traffic flow performance and safety.

\section*{CRediT authorship contribution statement}

\textbf{Peiyi Fang:} Methodology, Software, Data curation, Formal analysis, Investigation, Validation, Visualization, Writing -- original draft, Writing -- review and editing.

\textbf{Xiangyu Li:} Data curation, Resources, Writing -- review and editing.

\textbf{Yonglin Weng:} Data curation, Resources.

\textbf{Ke Ma:} Conceptualization, Methodology, Supervision, Project administration, Writing -- review and editing.

\bibliographystyle{elsarticle-num}
\bibliography{references}

\clearpage
\appendix
\setcounter{table}{0}
\renewcommand{\thetable}{\Alph{section}.\arabic{table}}

\section{Segment-Duration Statistics}
\label{app:duration_statistics}

Detailed duration statistics of the retained trajectory segments are reported below for the four scenario dimensions.

\begin{table}[!htb]
\centering
\caption{Duration statistics of trajectory segments by driving state.}
\label{tab:duration_driving_state}
\small
\setlength{\tabcolsep}{7pt}
\begin{tabular}{llccc}
\toprule
Driving state & Vehicle & Median duration (s) & \(Q_1\)--\(Q_3\) (s) & IQR (s) \\
\midrule
Cruise & HV   & 77.73 & 67.35--93.13  & 25.78 \\
& AV-1 & 76.36 & 66.48--91.83  & 25.35 \\
& AV-2 & 79.97 & 67.27--115.89 & 48.61 \\
& AV-3 & 74.00 & 70.11--106.64 & 36.53 \\
& AV-4 & 85.41 & 69.88--109.93 & 40.05 \\
\midrule
Follow & HV   & 80.05 & 67.93--98.91  & 30.98 \\
& AV-1 & 76.76 & 68.46--106.58 & 38.12 \\
& AV-2 & 99.15 & 76.09--146.61 & 70.52 \\
& AV-3 & 86.19 & 70.52--107.17 & 36.65 \\
& AV-4 & 89.91 & 69.95--125.54 & 55.59 \\
\midrule
Left turn & HV   & 7.75 & 6.78--8.75  & 1.97 \\
& AV-1 & 8.00 & 6.53--8.87  & 2.35 \\
& AV-2 & 9.92 & 9.23--10.16 & 0.93 \\
& AV-3 & 9.15 & 7.91--9.85  & 1.94 \\
& AV-4 & 8.95 & 7.83--10.01 & 2.18 \\
\midrule
Right turn & HV   & 7.52 & 6.66--8.50  & 1.85 \\
& AV-1 & 7.90 & 6.96--9.09  & 2.12 \\
& AV-2 & 8.85 & 8.59--10.48 & 1.89 \\
& AV-3 & 8.84 & 7.46--9.41  & 1.95 \\
& AV-4 & 8.26 & 7.60--9.03  & 1.43 \\
\bottomrule
\end{tabular}
\end{table}

\begin{table}[!htb]
\centering
\caption{Duration statistics of trajectory segments by lighting condition.}
\label{tab:duration_lighting}
\small
\setlength{\tabcolsep}{7pt}
\begin{tabular}{llccc}
\toprule
Lighting & Vehicle & Median duration (s) & \(Q_1\)--\(Q_3\) (s) & IQR (s) \\
\midrule
Day & HV   & 98.24  & 75.25--141.51 & 66.26 \\
& AV-1 & 86.27  & 71.45--117.45 & 46.00 \\
& AV-2 & 135.95 & 90.18--216.45 & 126.26 \\
& AV-3 & 87.95  & 71.75--115.80 & 44.06 \\
& AV-4 & 129.68 & 85.15--205.30 & 120.15 \\
\midrule
Night & HV   & 81.50  & 68.40--115.41 & 47.01 \\
& AV-1 & 91.73  & 71.50--119.37 & 47.87 \\
& AV-2 & 109.95 & 87.72--142.08 & 54.36 \\
& AV-3 & 85.11  & 74.55--127.91 & 53.36 \\
& AV-4 & 120.07 & 85.02--183.63 & 98.61 \\
\bottomrule
\end{tabular}
\end{table}

\begin{table}[!htb]
\centering
\caption{Duration statistics of trajectory segments by weather condition.}
\label{tab:duration_weather}
\small
\setlength{\tabcolsep}{7pt}
\begin{tabular}{llccc}
\toprule
Weather & Vehicle & Median duration (s) & \(Q_1\)--\(Q_3\) (s) & IQR (s) \\
\midrule
Clear & HV   & 89.90  & 68.75--123.79 & 55.04 \\
& AV-1 & 71.67  & 66.05--90.39  & 24.34 \\
& AV-2 & 104.50 & 76.93--168.43 & 91.51 \\
& AV-3 & 81.91  & 71.56--102.50 & 30.95 \\
& AV-4 & 86.85  & 70.20--120.48 & 50.28 \\
\midrule
Rain & HV   & 95.75  & 74.84--130.22 & 55.38 \\
& AV-1 & 81.56  & 65.73--95.93  & 30.20 \\
& AV-2 & 105.63 & 84.05--175.47 & 91.42 \\
& AV-3 & 90.22  & 76.78--120.77 & 43.99 \\
& AV-4 & 100.90 & 74.96--163.91 & 88.95 \\
\bottomrule
\end{tabular}
\end{table}

\begin{table}[!htb]
\centering
\caption{Duration statistics of trajectory segments by vehicle-density level.}
\label{tab:duration_density}
\small
\setlength{\tabcolsep}{7pt}
\begin{tabular}{llccc}
\toprule
Vehicle density & Vehicle & Median duration (s) & \(Q_1\)--\(Q_3\) (s) & IQR (s) \\
\midrule
High & HV   & 79.60 & 66.88--102.84 & 35.97 \\
& AV-1 & 83.30 & 69.94--120.94 & 51.00 \\
& AV-2 & 96.79 & 73.88--119.95 & 46.07 \\
& AV-3 & 77.92 & 65.03--95.37  & 30.34 \\
& AV-4 & 95.05 & 70.10--126.52 & 56.42 \\
\midrule
Low & HV   & 81.55  & 70.03--112.57 & 42.54 \\
& AV-1 & 82.11  & 71.24--99.32  & 28.07 \\
& AV-2 & 89.62  & 77.41--110.47 & 33.07 \\
& AV-3 & 85.38  & 70.24--118.49 & 48.25 \\
& AV-4 & 102.36 & 79.24--160.96 & 81.72 \\
\midrule
Medium & HV   & 73.80 & 66.84--89.88  & 23.04 \\
& AV-1 & 71.45 & 65.60--96.49  & 30.89 \\
& AV-2 & 96.81 & 77.19--145.23 & 68.04 \\
& AV-3 & 71.44 & 65.19--84.98  & 19.79 \\
& AV-4 & 84.91 & 69.90--104.51 & 34.61 \\
\bottomrule
\end{tabular}
\end{table}

\clearpage
\section{Complete Scenario--Signal--Feature Scores}
\label{app:complete_scores}

\begin{table}[!htb]
\centering
\scriptsize
\renewcommand{\arraystretch}{1.05}
\caption{Complete $S^{\mathrm{KDE}}$ values for the cruise condition across PAV platforms, kinematic signals, and frequency-domain features.}
\label{tab:appendix_kde_cruise}
\begin{tabular*}{\textwidth}{@{\extracolsep{\fill}}llccccccc@{}}
\toprule
Vehicle & Signal & Cen & Ent & F20 & F80 & Low & Mid & High \\
\midrule
\multirow{3}{*}{AV-1}
& Speed & 0.432 & 0.748 & 0.529 & 0.326 & 0.674 & 0.189 & 0.398 \\
& Acc   & 0.567 & 0.675 & 0.337 & 0.539 & 0.582 & 0.556 & 0.528 \\
& Jerk  & 0.690 & 0.776 & 0.612 & 0.649 & 0.670 & 0.687 & 0.820 \\
\midrule
\multirow{3}{*}{AV-2}
& Speed & 0.598 & 0.858 & 0.778 & 0.465 & 0.604 & 0.251 & 0.446 \\
& Acc   & 0.737 & 0.702 & 0.549 & 0.665 & 0.724 & 0.701 & 0.618 \\
& Jerk  & 0.727 & 0.797 & 0.615 & 0.787 & 0.683 & 0.689 & 0.692 \\
\midrule
\multirow{3}{*}{AV-3}
& Speed & 0.764 & 0.567 & 0.830 & 0.770 & 0.515 & 0.535 & 0.530 \\
& Acc   & 0.677 & 0.779 & 0.671 & 0.586 & 0.572 & 0.693 & 0.595 \\
& Jerk  & 0.708 & 0.735 & 0.593 & 0.646 & 0.672 & 0.679 & 0.816 \\
\midrule
\multirow{3}{*}{AV-4}
& Speed & 0.335 & 0.523 & 0.622 & 0.220 & 0.657 & 0.177 & 0.349 \\
& Acc   & 0.455 & 0.550 & 0.279 & 0.526 & 0.480 & 0.471 & 0.478 \\
& Jerk  & 0.676 & 0.783 & 0.614 & 0.692 & 0.629 & 0.650 & 0.837 \\
\bottomrule
\end{tabular*}
\end{table}

\begin{table}[!htb]
\centering
\scriptsize
\renewcommand{\arraystretch}{1.05}
\caption{Complete $S^{W}$ values for the cruise condition across PAV platforms, kinematic signals, and frequency-domain features.}
\label{tab:appendix_w_cruise}
\begin{tabular*}{\textwidth}{@{\extracolsep{\fill}}llccccccc@{}}
\toprule
Vehicle & Signal & Cen & Ent & F20 & F80 & Low & Mid & High \\
\midrule
\multirow{3}{*}{AV-1}
& Speed & 0.131 & 0.511 & 0.327 & 0.038 & 0.807 & 0.004 & 0.479 \\
& Acc   & 0.310 & 0.563 & 0.246 & 0.412 & 0.333 & 0.317 & 0.275 \\
& Jerk  & 0.657 & 0.771 & 0.580 & 0.715 & 0.723 & 0.726 & 0.691 \\
\midrule
\multirow{3}{*}{AV-2}
& Speed & 0.378 & 0.718 & 0.582 & 0.207 & 0.632 & 0.100 & 0.669 \\
& Acc   & 0.634 & 0.533 & 0.654 & 0.691 & 0.645 & 0.676 & 0.460 \\
& Jerk  & 0.755 & 0.875 & 0.727 & 0.858 & 0.734 & 0.704 & 0.547 \\
\midrule
\multirow{3}{*}{AV-3}
& Speed & 0.593 & 0.276 & 0.606 & 0.573 & 0.606 & 0.576 & 0.679 \\
& Acc   & 0.657 & 0.736 & 0.690 & 0.592 & 0.497 & 0.713 & 0.525 \\
& Jerk  & 0.792 & 0.667 & 0.705 & 0.848 & 0.752 & 0.746 & 0.787 \\
\midrule
\multirow{3}{*}{AV-4}
& Speed & 0.012 & 0.138 & 0.250 & $1.17\times10^{-4}$ & 0.670 & $5.42\times10^{-5}$ & 0.279 \\
& Acc   & 0.170 & 0.359 & 0.067 & 0.323 & 0.193 & 0.184 & 0.145 \\
& Jerk  & 0.617 & 0.806 & 0.535 & 0.685 & 0.651 & 0.654 & 0.715 \\
\bottomrule
\end{tabular*}
\end{table}

\begin{table}[!htb]
\centering
\scriptsize
\renewcommand{\arraystretch}{1.05}
\caption{Complete $S^{\mathrm{KDE}}$ values for the car-following condition across PAV platforms, kinematic signals, and frequency-domain features.}
\label{tab:appendix_kde_follow}
\begin{tabular*}{\textwidth}{@{\extracolsep{\fill}}llccccccc@{}}
\toprule
Vehicle & Signal & Cen & Ent & F20 & F80 & Low & Mid & High \\
\midrule
\multirow{3}{*}{AV-1}
& Speed & 0.786 & 0.741 & 0.863 & 0.737 & 0.633 & 0.495 & 0.603 \\
& Acc   & 0.628 & 0.775 & 0.686 & 0.598 & 0.621 & 0.612 & 0.651 \\
& Jerk  & 0.703 & 0.778 & 0.665 & 0.709 & 0.743 & 0.731 & 0.816 \\
\midrule
\multirow{3}{*}{AV-2}
& Speed & 0.673 & 0.834 & 0.776 & 0.701 & 0.624 & 0.468 & 0.581 \\
& Acc   & 0.679 & 0.755 & 0.719 & 0.608 & 0.618 & 0.674 & 0.564 \\
& Jerk  & 0.739 & 0.867 & 0.639 & 0.751 & 0.778 & 0.821 & 0.633 \\
\midrule
\multirow{3}{*}{AV-3}
& Speed & 0.751 & 0.748 & 0.828 & 0.723 & 0.657 & 0.708 & 0.463 \\
& Acc   & 0.696 & 0.739 & 0.660 & 0.668 & 0.745 & 0.690 & 0.809 \\
& Jerk  & 0.734 & 0.742 & 0.634 & 0.719 & 0.697 & 0.686 & 0.756 \\
\midrule
\multirow{3}{*}{AV-4}
& Speed & 0.695 & 0.748 & 0.800 & 0.642 & 0.638 & 0.305 & 0.490 \\
& Acc   & 0.625 & 0.735 & 0.609 & 0.542 & 0.546 & 0.567 & 0.547 \\
& Jerk  & 0.649 & 0.805 & 0.553 & 0.685 & 0.654 & 0.675 & 0.788 \\
\bottomrule
\end{tabular*}
\end{table}

\begin{table}[!htb]
\centering
\scriptsize
\renewcommand{\arraystretch}{1.05}
\caption{Complete $S^{W}$ values for the car-following condition across PAV platforms, kinematic signals, and frequency-domain features.}
\label{tab:appendix_w_follow}
\begin{tabular*}{\textwidth}{@{\extracolsep{\fill}}llccccccc@{}}
\toprule
Vehicle & Signal & Cen & Ent & F20 & F80 & Low & Mid & High \\
\midrule
\multirow{3}{*}{AV-1}
& Speed & 0.530 & 0.541 & 0.742 & 0.490 & 0.816 & 0.473 & 0.832 \\
& Acc   & 0.497 & 0.684 & 0.648 & 0.446 & 0.488 & 0.495 & 0.506 \\
& Jerk  & 0.677 & 0.736 & 0.707 & 0.708 & 0.760 & 0.761 & 0.858 \\
\midrule
\multirow{3}{*}{AV-2}
& Speed & 0.406 & 0.594 & 0.601 & 0.380 & 0.522 & 0.264 & 0.839 \\
& Acc   & 0.521 & 0.611 & 0.584 & 0.468 & 0.395 & 0.543 & 0.233 \\
& Jerk  & 0.672 & 0.851 & 0.607 & 0.750 & 0.796 & 0.867 & 0.389 \\
\midrule
\multirow{3}{*}{AV-3}
& Speed & 0.579 & 0.515 & 0.679 & 0.535 & 0.684 & 0.662 & 0.576 \\
& Acc   & 0.635 & 0.542 & 0.665 & 0.621 & 0.666 & 0.663 & 0.762 \\
& Jerk  & 0.664 & 0.613 & 0.698 & 0.669 & 0.614 & 0.600 & 0.752 \\
\midrule
\multirow{3}{*}{AV-4}
& Speed & 0.339 & 0.543 & 0.604 & 0.259 & 0.580 & 0.078 & 0.740 \\
& Acc   & 0.366 & 0.550 & 0.469 & 0.300 & 0.275 & 0.307 & 0.215 \\
& Jerk  & 0.515 & 0.781 & 0.412 & 0.620 & 0.548 & 0.568 & 0.613 \\
\bottomrule
\end{tabular*}
\end{table}

\begin{table}[!htb]
\centering
\scriptsize
\renewcommand{\arraystretch}{1.05}
\caption{Complete $S^{\mathrm{KDE}}$ values for the left-turn condition across PAV platforms, kinematic signals, and frequency-domain features.}
\label{tab:appendix_kde_left}
\begin{tabular*}{\textwidth}{@{\extracolsep{\fill}}llccccccc@{}}
\toprule
Vehicle & Signal & Cen & Ent & F20 & F80 & Low & Mid & High \\
\midrule
\multirow{3}{*}{AV-1}
& Yaw rate & 0.684 & 0.530 & 0.870 & 0.600 & 0.157 & 0.158 & 0.737 \\
& Yaw acc  & 0.644 & 0.487 & 0.759 & 0.475 & 0.448 & 0.401 & 0.584 \\
& Yaw jerk & 0.439 & 0.550 & 0.632 & 0.430 & 0.832 & 0.707 & 0.406 \\
\midrule
\multirow{3}{*}{AV-2}
& Yaw rate & 0.637 & 0.765 & 0.400 & 0.730 & 0.570 & 0.571 & 0.673 \\
& Yaw acc  & 0.573 & 0.756 & 0.604 & 0.649 & 0.717 & 0.710 & 0.820 \\
& Yaw jerk & 0.711 & 0.669 & 0.627 & 0.808 & 0.831 & 0.761 & 0.661 \\
\midrule
\multirow{3}{*}{AV-3}
& Yaw rate & 0.558 & 0.854 & 0.707 & 0.641 & 0.553 & 0.553 & 0.704 \\
& Yaw acc  & 0.532 & 0.723 & 0.648 & 0.575 & 0.613 & 0.614 & 0.860 \\
& Yaw jerk & 0.829 & 0.736 & 0.640 & 0.773 & 0.706 & 0.638 & 0.734 \\
\midrule
\multirow{3}{*}{AV-4}
& Yaw rate & 0.675 & 0.833 & 0.766 & 0.728 & 0.454 & 0.451 & 0.843 \\
& Yaw acc  & 0.741 & 0.797 & 0.734 & 0.713 & 0.741 & 0.719 & 0.652 \\
& Yaw jerk & 0.681 & 0.621 & 0.781 & 0.619 & 0.896 & 0.749 & 0.568 \\
\bottomrule
\end{tabular*}
\end{table}

\begin{table}[!htb]
\centering
\scriptsize
\renewcommand{\arraystretch}{1.05}
\caption{Complete $S^{W}$ values for the left-turn condition across PAV platforms, kinematic signals, and frequency-domain features.}
\label{tab:appendix_w_left}
\begin{tabular*}{\textwidth}{@{\extracolsep{\fill}}llccccccc@{}}
\toprule
Vehicle & Signal & Cen & Ent & F20 & F80 & Low & Mid & High \\
\midrule
\multirow{3}{*}{AV-1}
& Yaw rate & 0.500 & 0.345 & 0.776 & 0.499 & 0.535 & 0.535 & 0.729 \\
& Yaw acc  & 0.392 & 0.269 & 0.614 & 0.284 & 0.354 & 0.340 & 0.405 \\
& Yaw jerk & 0.145 & 0.279 & 0.466 & 0.148 & 0.784 & 0.467 & 0.142 \\
\midrule
\multirow{3}{*}{AV-2}
& Yaw rate & 0.419 & 0.663 & 0.324 & 0.567 & 0.657 & 0.658 & 0.682 \\
& Yaw acc  & 0.372 & 0.700 & 0.428 & 0.475 & 0.562 & 0.561 & 0.781 \\
& Yaw jerk & 0.605 & 0.441 & 0.470 & 0.687 & 0.717 & 0.620 & 0.625 \\
\midrule
\multirow{3}{*}{AV-3}
& Yaw rate & 0.402 & 0.750 & 0.505 & 0.492 & 0.655 & 0.656 & 0.695 \\
& Yaw acc  & 0.344 & 0.556 & 0.482 & 0.371 & 0.460 & 0.463 & 0.811 \\
& Yaw jerk & 0.789 & 0.496 & 0.488 & 0.587 & 0.531 & 0.444 & 0.559 \\
\midrule
\multirow{3}{*}{AV-4}
& Yaw rate & 0.462 & 0.686 & 0.586 & 0.528 & 0.622 & 0.622 & 0.835 \\
& Yaw acc  & 0.522 & 0.653 & 0.548 & 0.509 & 0.559 & 0.540 & 0.465 \\
& Yaw jerk & 0.406 & 0.395 & 0.675 & 0.325 & 0.863 & 0.614 & 0.319 \\
\bottomrule
\end{tabular*}
\end{table}

\begin{table}[!htb]
\centering
\scriptsize
\renewcommand{\arraystretch}{1.05}
\caption{Complete $S^{\mathrm{KDE}}$ values for the right-turn condition across PAV platforms, kinematic signals, and frequency-domain features.}
\label{tab:appendix_kde_right}
\begin{tabular*}{\textwidth}{@{\extracolsep{\fill}}llccccccc@{}}
\toprule
Vehicle & Signal & Cen & Ent & F20 & F80 & Low & Mid & High \\
\midrule
\multirow{3}{*}{AV-1}
& Yaw rate & 0.728 & 0.804 & 0.865 & 0.786 & 0.419 & 0.415 & 0.672 \\
& Yaw acc  & 0.657 & 0.657 & 0.796 & 0.682 & 0.679 & 0.663 & 0.520 \\
& Yaw jerk & 0.756 & 0.682 & 0.701 & 0.629 & 0.826 & 0.785 & 0.596 \\
\midrule
\multirow{3}{*}{AV-2}
& Yaw rate & 0.723 & 0.742 & 0.629 & 0.776 & 0.579 & 0.578 & 0.858 \\
& Yaw acc  & 0.708 & 0.895 & 0.635 & 0.790 & 0.701 & 0.699 & 0.718 \\
& Yaw jerk & 0.818 & 0.872 & 0.631 & 0.825 & 0.729 & 0.772 & 0.712 \\
\midrule
\multirow{3}{*}{AV-3}
& Yaw rate & 0.679 & 0.793 & 0.761 & 0.705 & 0.627 & 0.625 & 0.472 \\
& Yaw acc  & 0.714 & 0.744 & 0.732 & 0.726 & 0.749 & 0.751 & 0.598 \\
& Yaw jerk & 0.670 & 0.717 & 0.756 & 0.680 & 0.894 & 0.824 & 0.686 \\
\midrule
\multirow{3}{*}{AV-4}
& Yaw rate & 0.632 & 0.791 & 0.748 & 0.717 & 0.407 & 0.404 & 0.816 \\
& Yaw acc  & 0.818 & 0.874 & 0.781 & 0.784 & 0.792 & 0.770 & 0.776 \\
& Yaw jerk & 0.660 & 0.677 & 0.785 & 0.723 & 0.898 & 0.858 & 0.724 \\
\bottomrule
\end{tabular*}
\end{table}

\begin{table}[!htb]
\centering
\scriptsize
\renewcommand{\arraystretch}{1.05}
\caption{Complete $S^{W}$ values for the right-turn condition across PAV platforms, kinematic signals, and frequency-domain features.}
\label{tab:appendix_w_right}
\begin{tabular*}{\textwidth}{@{\extracolsep{\fill}}llccccccc@{}}
\toprule
Vehicle & Signal & Cen & Ent & F20 & F80 & Low & Mid & High \\
\midrule
\multirow{3}{*}{AV-1}
& Yaw rate & 0.505 & 0.625 & 0.649 & 0.607 & 0.634 & 0.632 & 0.723 \\
& Yaw acc  & 0.408 & 0.434 & 0.619 & 0.451 & 0.518 & 0.507 & 0.391 \\
& Yaw jerk & 0.412 & 0.394 & 0.632 & 0.368 & 0.775 & 0.574 & 0.309 \\
\midrule
\multirow{3}{*}{AV-2}
& Yaw rate & 0.569 & 0.558 & 0.417 & 0.684 & 0.707 & 0.707 & 0.817 \\
& Yaw acc  & 0.498 & 0.781 & 0.450 & 0.620 & 0.515 & 0.521 & 0.642 \\
& Yaw jerk & 0.632 & 0.722 & 0.521 & 0.699 & 0.475 & 0.514 & 0.663 \\
\midrule
\multirow{3}{*}{AV-3}
& Yaw rate & 0.479 & 0.698 & 0.528 & 0.509 & 0.752 & 0.750 & 0.538 \\
& Yaw acc  & 0.578 & 0.621 & 0.595 & 0.561 & 0.679 & 0.675 & 0.452 \\
& Yaw jerk & 0.507 & 0.485 & 0.696 & 0.512 & 0.853 & 0.671 & 0.368 \\
\midrule
\multirow{3}{*}{AV-4}
& Yaw rate & 0.471 & 0.641 & 0.576 & 0.546 & 0.640 & 0.640 & 0.815 \\
& Yaw acc  & 0.631 & 0.776 & 0.616 & 0.626 & 0.601 & 0.590 & 0.730 \\
& Yaw jerk & 0.420 & 0.465 & 0.733 & 0.483 & 0.887 & 0.755 & 0.446 \\
\bottomrule
\end{tabular*}
\end{table}

\begin{table}[!htb]
\centering
\scriptsize
\renewcommand{\arraystretch}{1.05}
\caption{Complete $S^{\mathrm{KDE}}$ values for the day condition across PAV platforms, kinematic signals, and frequency-domain features.}
\label{tab:appendix_kde_day}
\begin{tabular*}{\textwidth}{@{\extracolsep{\fill}}llccccccc@{}}
\toprule
Vehicle & Signal & Cen & Ent & F20 & F80 & Low & Mid & High \\
\midrule
\multirow{3}{*}{AV-1}
& Speed & 0.673 & 0.883 & 0.899 & 0.579 & 0.759 & 0.204 & 0.355 \\
& Acc   & 0.660 & 0.814 & 0.498 & 0.598 & 0.788 & 0.610 & 0.633 \\
& Jerk  & 0.818 & 0.855 & 0.678 & 0.822 & 0.835 & 0.813 & 0.865 \\
\midrule
\multirow{3}{*}{AV-2}
& Speed & 0.798 & 0.936 & 0.818 & 0.781 & 0.729 & 0.341 & 0.515 \\
& Acc   & 0.720 & 0.858 & 0.613 & 0.657 & 0.750 & 0.683 & 0.558 \\
& Jerk  & 0.814 & 0.858 & 0.688 & 0.792 & 0.852 & 0.862 & 0.696 \\
\midrule
\multirow{3}{*}{AV-3}
& Speed & 0.917 & 0.892 & 0.856 & 0.913 & 0.763 & 0.544 & 0.452 \\
& Acc   & 0.844 & 0.753 & 0.912 & 0.794 & 0.868 & 0.805 & 0.816 \\
& Jerk  & 0.760 & 0.766 & 0.682 & 0.736 & 0.753 & 0.729 & 0.783 \\
\midrule
\multirow{3}{*}{AV-4}
& Speed & 0.722 & 0.867 & 0.816 & 0.653 & 0.714 & 0.281 & 0.498 \\
& Acc   & 0.683 & 0.806 & 0.587 & 0.606 & 0.643 & 0.624 & 0.647 \\
& Jerk  & 0.735 & 0.851 & 0.641 & 0.760 & 0.740 & 0.727 & 0.787 \\
\bottomrule
\end{tabular*}
\end{table}

\begin{table}[!htb]
\centering
\scriptsize
\renewcommand{\arraystretch}{1.05}
\caption{Complete $S^{W}$ values for the day condition across PAV platforms, kinematic signals, and frequency-domain features.}
\label{tab:appendix_w_day}
\begin{tabular*}{\textwidth}{@{\extracolsep{\fill}}llccccccc@{}}
\toprule
Vehicle & Signal & Cen & Ent & F20 & F80 & Low & Mid & High \\
\midrule
\multirow{3}{*}{AV-1}
& Speed & 0.428 & 0.716 & 0.754 & 0.356 & 0.943 & 0.039 & 0.417 \\
& Acc   & 0.381 & 0.722 & 0.205 & 0.429 & 0.638 & 0.349 & 0.475 \\
& Jerk  & 0.830 & 0.850 & 0.794 & 0.893 & 0.874 & 0.875 & 0.904 \\
\midrule
\multirow{3}{*}{AV-2}
& Speed & 0.544 & 0.860 & 0.688 & 0.479 & 0.693 & 0.261 & 0.770 \\
& Acc   & 0.464 & 0.741 & 0.314 & 0.539 & 0.529 & 0.467 & 0.343 \\
& Jerk  & 0.810 & 0.837 & 0.765 & 0.851 & 0.882 & 0.935 & 0.506 \\
\midrule
\multirow{3}{*}{AV-3}
& Speed & 0.793 & 0.770 & 0.785 & 0.854 & 0.881 & 0.585 & 0.555 \\
& Acc   & 0.740 & 0.613 & 0.815 & 0.747 & 0.829 & 0.718 & 0.778 \\
& Jerk  & 0.731 & 0.646 & 0.743 & 0.729 & 0.669 & 0.683 & 0.803 \\
\midrule
\multirow{3}{*}{AV-4}
& Speed & 0.365 & 0.706 & 0.682 & 0.254 & 0.679 & 0.063 & 0.630 \\
& Acc   & 0.332 & 0.530 & 0.250 & 0.361 & 0.366 & 0.269 & 0.401 \\
& Jerk  & 0.590 & 0.744 & 0.486 & 0.682 & 0.583 & 0.605 & 0.730 \\
\bottomrule
\end{tabular*}
\end{table}

\begin{table}[!htb]
\centering
\scriptsize
\renewcommand{\arraystretch}{1.05}
\caption{Complete $S^{\mathrm{KDE}}$ values for the night-lit condition across PAV platforms, kinematic signals, and frequency-domain features.}
\label{tab:appendix_kde_night_lit}
\begin{tabular*}{\textwidth}{@{\extracolsep{\fill}}llccccccc@{}}
\toprule
Vehicle & Signal & Cen & Ent & F20 & F80 & Low & Mid & High \\
\midrule
\multirow{3}{*}{AV-1}
& Speed & 0.692 & 0.808 & 0.782 & 0.660 & 0.660 & 0.273 & 0.529 \\
& Acc   & 0.614 & 0.731 & 0.363 & 0.580 & 0.670 & 0.573 & 0.597 \\
& Jerk  & 0.771 & 0.774 & 0.637 & 0.754 & 0.727 & 0.709 & 0.794 \\
\midrule
\multirow{3}{*}{AV-2}
& Speed & 0.593 & 0.652 & 0.685 & 0.572 & 0.560 & 0.343 & 0.306 \\
& Acc   & 0.678 & 0.649 & 0.662 & 0.645 & 0.666 & 0.691 & 0.821 \\
& Jerk  & 0.732 & 0.684 & 0.586 & 0.666 & 0.683 & 0.697 & 0.581 \\
\midrule
\multirow{3}{*}{AV-3}
& Speed & 0.832 & 0.794 & 0.859 & 0.783 & 0.630 & 0.725 & 0.523 \\
& Acc   & 0.676 & 0.709 & 0.723 & 0.647 & 0.805 & 0.618 & 0.855 \\
& Jerk  & 0.600 & 0.713 & 0.582 & 0.596 & 0.588 & 0.566 & 0.659 \\
\midrule
\multirow{3}{*}{AV-4}
& Speed & 0.700 & 0.781 & 0.720 & 0.684 & 0.545 & 0.629 & 0.417 \\
& Acc   & 0.878 & 0.848 & 0.724 & 0.765 & 0.676 & 0.818 & 0.743 \\
& Jerk  & 0.823 & 0.797 & 0.754 & 0.751 & 0.784 & 0.761 & 0.794 \\
\bottomrule
\end{tabular*}
\end{table}

\begin{table}[!htb]
\centering
\scriptsize
\renewcommand{\arraystretch}{1.05}
\caption{Complete $S^{W}$ values for the night-lit condition across PAV platforms, kinematic signals, and frequency-domain features.}
\label{tab:appendix_w_night_lit}
\begin{tabular*}{\textwidth}{@{\extracolsep{\fill}}llccccccc@{}}
\toprule
Vehicle & Signal & Cen & Ent & F20 & F80 & Low & Mid & High \\
\midrule
\multirow{3}{*}{AV-1}
& Speed & 0.470 & 0.676 & 0.634 & 0.423 & 0.687 & 0.094 & 0.733 \\
& Acc   & 0.318 & 0.622 & 0.039 & 0.451 & 0.358 & 0.298 & 0.342 \\
& Jerk  & 0.815 & 0.795 & 0.727 & 0.850 & 0.845 & 0.838 & 0.819 \\
\midrule
\multirow{3}{*}{AV-2}
& Speed & 0.418 & 0.487 & 0.461 & 0.460 & 0.443 & 0.540 & 0.689 \\
& Acc   & 0.546 & 0.502 & 0.492 & 0.591 & 0.604 & 0.604 & 0.761 \\
& Jerk  & 0.675 & 0.625 & 0.596 & 0.686 & 0.707 & 0.740 & 0.317 \\
\midrule
\multirow{3}{*}{AV-3}
& Speed & 0.758 & 0.680 & 0.741 & 0.692 & 0.715 & 0.885 & 0.769 \\
& Acc   & 0.620 & 0.535 & 0.623 & 0.653 & 0.855 & 0.608 & 0.864 \\
& Jerk  & 0.573 & 0.625 & 0.642 & 0.586 & 0.530 & 0.531 & 0.690 \\
\midrule
\multirow{3}{*}{AV-4}
& Speed & 0.521 & 0.708 & 0.524 & 0.540 & 0.472 & 0.667 & 0.772 \\
& Acc   & 0.847 & 0.764 & 0.580 & 0.824 & 0.482 & 0.830 & 0.616 \\
& Jerk  & 0.839 & 0.759 & 0.906 & 0.809 & 0.858 & 0.875 & 0.722 \\
\bottomrule
\end{tabular*}
\end{table}

\begin{table}[!htb]
\centering
\scriptsize
\renewcommand{\arraystretch}{1.05}
\caption{Complete $S^{\mathrm{KDE}}$ values for the clear-weather condition across PAV platforms, kinematic signals, and frequency-domain features.}
\label{tab:appendix_kde_clear}
\begin{tabular*}{\textwidth}{@{\extracolsep{\fill}}llccccccc@{}}
\toprule
Vehicle & Signal & Cen & Ent & F20 & F80 & Low & Mid & High \\
\midrule
\multirow{3}{*}{AV-1}
& Speed & 0.915 & 0.723 & 0.801 & 0.819 & 0.650 & 0.721 & 0.659 \\
& Acc   & 0.685 & 0.725 & 0.890 & 0.614 & 0.712 & 0.621 & 0.808 \\
& Jerk  & 0.729 & 0.710 & 0.621 & 0.663 & 0.685 & 0.675 & 0.620 \\
\midrule
\multirow{3}{*}{AV-2}
& Speed & 0.764 & 0.841 & 0.857 & 0.774 & 0.677 & 0.625 & 0.564 \\
& Acc   & 0.825 & 0.811 & 0.857 & 0.746 & 0.855 & 0.787 & 0.599 \\
& Jerk  & 0.833 & 0.813 & 0.746 & 0.786 & 0.843 & 0.816 & 0.670 \\
\midrule
\multirow{3}{*}{AV-3}
& Speed & 0.867 & 0.782 & 0.918 & 0.805 & 0.723 & 0.679 & 0.489 \\
& Acc   & 0.843 & 0.745 & 0.831 & 0.763 & 0.835 & 0.799 & 0.763 \\
& Jerk  & 0.752 & 0.818 & 0.682 & 0.671 & 0.725 & 0.705 & 0.754 \\
\midrule
\multirow{3}{*}{AV-4}
& Speed & 0.755 & 0.804 & 0.913 & 0.674 & 0.705 & 0.266 & 0.563 \\
& Acc   & 0.725 & 0.900 & 0.567 & 0.651 & 0.801 & 0.663 & 0.676 \\
& Jerk  & 0.796 & 0.886 & 0.717 & 0.791 & 0.817 & 0.777 & 0.839 \\
\bottomrule
\end{tabular*}
\end{table}

\begin{table}[!htb]
\centering
\scriptsize
\renewcommand{\arraystretch}{1.05}
\caption{Complete $S^{W}$ values for the clear-weather condition across PAV platforms, kinematic signals, and frequency-domain features.}
\label{tab:appendix_w_clear}
\begin{tabular*}{\textwidth}{@{\extracolsep{\fill}}llccccccc@{}}
\toprule
Vehicle & Signal & Cen & Ent & F20 & F80 & Low & Mid & High \\
\midrule
\multirow{3}{*}{AV-1}
& Speed & 0.826 & 0.498 & 0.699 & 0.704 & 0.779 & 0.697 & 0.768 \\
& Acc   & 0.681 & 0.580 & 0.840 & 0.647 & 0.677 & 0.659 & 0.835 \\
& Jerk  & 0.721 & 0.553 & 0.736 & 0.699 & 0.661 & 0.697 & 0.560 \\
\midrule
\multirow{3}{*}{AV-2}
& Speed & 0.587 & 0.706 & 0.651 & 0.639 & 0.709 & 0.591 & 0.835 \\
& Acc   & 0.803 & 0.711 & 0.719 & 0.752 & 0.780 & 0.815 & 0.456 \\
& Jerk  & 0.825 & 0.723 & 0.895 & 0.791 & 0.906 & 0.876 & 0.473 \\
\midrule
\multirow{3}{*}{AV-3}
& Speed & 0.825 & 0.627 & 0.882 & 0.751 & 0.925 & 0.699 & 0.604 \\
& Acc   & 0.838 & 0.666 & 0.763 & 0.806 & 0.851 & 0.849 & 0.691 \\
& Jerk  & 0.792 & 0.795 & 0.782 & 0.780 & 0.715 & 0.738 & 0.817 \\
\midrule
\multirow{3}{*}{AV-4}
& Speed & 0.480 & 0.635 & 0.809 & 0.382 & 0.874 & 0.093 & 0.776 \\
& Acc   & 0.592 & 0.851 & 0.339 & 0.590 & 0.624 & 0.560 & 0.479 \\
& Jerk  & 0.834 & 0.913 & 0.783 & 0.900 & 0.862 & 0.868 & 0.904 \\
\bottomrule
\end{tabular*}
\end{table}

\begin{table}[!htb]
\centering
\scriptsize
\renewcommand{\arraystretch}{1.05}
\caption{Complete $S^{\mathrm{KDE}}$ values for the rain condition across PAV platforms, kinematic signals, and frequency-domain features.}
\label{tab:appendix_kde_rain}
\begin{tabular*}{\textwidth}{@{\extracolsep{\fill}}llccccccc@{}}
\toprule
Vehicle & Signal & Cen & Ent & F20 & F80 & Low & Mid & High \\
\midrule
\multirow{3}{*}{AV-1}
& Speed & 0.589 & 0.818 & 0.821 & 0.515 & 0.730 & 0.190 & 0.338 \\
& Acc   & 0.575 & 0.767 & 0.418 & 0.534 & 0.822 & 0.528 & 0.581 \\
& Jerk  & 0.730 & 0.757 & 0.616 & 0.746 & 0.751 & 0.730 & 0.807 \\
\midrule
\multirow{3}{*}{AV-2}
& Speed & 0.780 & 0.788 & 0.831 & 0.760 & 0.675 & 0.533 & 0.557 \\
& Acc   & 0.711 & 0.843 & 0.611 & 0.650 & 0.702 & 0.681 & 0.417 \\
& Jerk  & 0.779 & 0.845 & 0.648 & 0.783 & 0.782 & 0.801 & 0.630 \\
\midrule
\multirow{3}{*}{AV-3}
& Speed & 0.897 & 0.846 & 0.896 & 0.883 & 0.715 & 0.695 & 0.544 \\
& Acc   & 0.798 & 0.703 & 0.902 & 0.717 & 0.814 & 0.720 & 0.841 \\
& Jerk  & 0.715 & 0.682 & 0.607 & 0.701 & 0.653 & 0.623 & 0.733 \\
\midrule
\multirow{3}{*}{AV-4}
& Speed & 0.608 & 0.759 & 0.764 & 0.469 & 0.708 & 0.204 & 0.383 \\
& Acc   & 0.640 & 0.776 & 0.459 & 0.614 & 0.680 & 0.584 & 0.590 \\
& Jerk  & 0.712 & 0.827 & 0.626 & 0.736 & 0.691 & 0.690 & 0.750 \\
\bottomrule
\end{tabular*}
\end{table}

\begin{table}[!htb]
\centering
\scriptsize
\renewcommand{\arraystretch}{1.05}
\caption{Complete $S^{W}$ values for the rain condition across PAV platforms, kinematic signals, and frequency-domain features.}
\label{tab:appendix_w_rain}
\begin{tabular*}{\textwidth}{@{\extracolsep{\fill}}llccccccc@{}}
\toprule
Vehicle & Signal & Cen & Ent & F20 & F80 & Low & Mid & High \\
\midrule
\multirow{3}{*}{AV-1}
& Speed & 0.373 & 0.657 & 0.606 & 0.358 & 0.891 & 0.052 & 0.332 \\
& Acc   & 0.302 & 0.623 & 0.131 & 0.373 & 0.645 & 0.285 & 0.476 \\
& Jerk  & 0.739 & 0.650 & 0.673 & 0.799 & 0.789 & 0.780 & 0.848 \\
\midrule
\multirow{3}{*}{AV-2}
& Speed & 0.611 & 0.669 & 0.630 & 0.598 & 0.644 & 0.622 & 0.840 \\
& Acc   & 0.565 & 0.769 & 0.438 & 0.583 & 0.596 & 0.599 & 0.294 \\
& Jerk  & 0.791 & 0.835 & 0.733 & 0.855 & 0.850 & 0.899 & 0.489 \\
\midrule
\multirow{3}{*}{AV-3}
& Speed & 0.861 & 0.766 & 0.862 & 0.862 & 0.872 & 0.851 & 0.733 \\
& Acc   & 0.678 & 0.482 & 0.812 & 0.675 & 0.805 & 0.646 & 0.876 \\
& Jerk  & 0.652 & 0.625 & 0.621 & 0.726 & 0.580 & 0.564 & 0.779 \\
\midrule
\multirow{3}{*}{AV-4}
& Speed & 0.188 & 0.519 & 0.582 & 0.081 & 0.741 & 0.010 & 0.370 \\
& Acc   & 0.273 & 0.494 & 0.117 & 0.390 & 0.417 & 0.223 & 0.369 \\
& Jerk  & 0.572 & 0.786 & 0.498 & 0.663 & 0.582 & 0.591 & 0.684 \\
\bottomrule
\end{tabular*}
\end{table}

\begin{table}[!htb]
\centering
\scriptsize
\renewcommand{\arraystretch}{1.05}
\caption{Complete $S^{\mathrm{KDE}}$ values for the low-density condition across PAV platforms, kinematic signals, and frequency-domain features.}
\label{tab:appendix_kde_density_low}
\begin{tabular*}{\textwidth}{@{\extracolsep{\fill}}llccccccc@{}}
\toprule
Vehicle & Signal & Cen & Ent & F20 & F80 & Low & Mid & High \\
\midrule
\multirow{3}{*}{AV-1}
& Speed & 0.634 & 0.785 & 0.724 & 0.584 & 0.672 & 0.215 & 0.459 \\
& Acc   & 0.514 & 0.641 & 0.295 & 0.485 & 0.608 & 0.468 & 0.458 \\
& Jerk  & 0.735 & 0.739 & 0.598 & 0.721 & 0.704 & 0.696 & 0.797 \\
\midrule
\multirow{3}{*}{AV-2}
& Speed & 0.479 & 0.686 & 0.827 & 0.310 & 0.579 & 0.175 & 0.297 \\
& Acc   & 0.521 & 0.738 & 0.230 & 0.567 & 0.618 & 0.491 & 0.287 \\
& Jerk  & 0.681 & 0.791 & 0.643 & 0.648 & 0.659 & 0.665 & 0.578 \\
\midrule
\multirow{3}{*}{AV-3}
& Speed & 0.781 & 0.878 & 0.898 & 0.815 & 0.646 & 0.440 & 0.323 \\
& Acc   & 0.773 & 0.783 & 0.791 & 0.685 & 0.714 & 0.735 & 0.717 \\
& Jerk  & 0.660 & 0.717 & 0.642 & 0.649 & 0.654 & 0.642 & 0.689 \\
\midrule
\multirow{3}{*}{AV-4}
& Speed & 0.606 & 0.755 & 0.805 & 0.475 & 0.646 & 0.247 & 0.454 \\
& Acc   & 0.589 & 0.665 & 0.388 & 0.532 & 0.621 & 0.550 & 0.448 \\
& Jerk  & 0.770 & 0.775 & 0.685 & 0.765 & 0.731 & 0.733 & 0.756 \\
\bottomrule
\end{tabular*}
\end{table}

\begin{table}[!htb]
\centering
\scriptsize
\renewcommand{\arraystretch}{1.05}
\caption{Complete $S^{W}$ values for the low-density condition across PAV platforms, kinematic signals, and frequency-domain features.}
\label{tab:appendix_w_density_low}
\begin{tabular*}{\textwidth}{@{\extracolsep{\fill}}llccccccc@{}}
\toprule
Vehicle & Signal & Cen & Ent & F20 & F80 & Low & Mid & High \\
\midrule
\multirow{3}{*}{AV-1}
& Speed & 0.393 & 0.593 & 0.546 & 0.362 & 0.816 & 0.026 & 0.579 \\
& Acc   & 0.154 & 0.401 & 0.006 & 0.261 & 0.311 & 0.143 & 0.123 \\
& Jerk  & 0.641 & 0.737 & 0.551 & 0.700 & 0.683 & 0.671 & 0.797 \\
\midrule
\multirow{3}{*}{AV-2}
& Speed & 0.159 & 0.476 & 0.754 & 0.032 & 0.539 & 0.003 & 0.169 \\
& Acc   & 0.281 & 0.582 & 0.006 & 0.561 & 0.384 & 0.301 & 0.029 \\
& Jerk  & 0.682 & 0.773 & 0.752 & 0.687 & 0.704 & 0.726 & 0.533 \\
\midrule
\multirow{3}{*}{AV-3}
& Speed & 0.682 & 0.803 & 0.840 & 0.751 & 0.840 & 0.402 & 0.294 \\
& Acc   & 0.728 & 0.633 & 0.609 & 0.752 & 0.738 & 0.741 & 0.691 \\
& Jerk  & 0.632 & 0.569 & 0.712 & 0.707 & 0.645 & 0.627 & 0.660 \\
\midrule
\multirow{3}{*}{AV-4}
& Speed & 0.180 & 0.386 & 0.669 & 0.073 & 0.652 & 0.006 & 0.413 \\
& Acc   & 0.227 & 0.358 & 0.031 & 0.327 & 0.320 & 0.222 & 0.083 \\
& Jerk  & 0.648 & 0.733 & 0.657 & 0.677 & 0.668 & 0.693 & 0.571 \\
\bottomrule
\end{tabular*}
\end{table}

\begin{table}[!htb]
\centering
\scriptsize
\renewcommand{\arraystretch}{1.05}
\caption{Complete $S^{\mathrm{KDE}}$ values for the medium-density condition across PAV platforms, kinematic signals, and frequency-domain features.}
\label{tab:appendix_kde_density_medium}
\begin{tabular*}{\textwidth}{@{\extracolsep{\fill}}llccccccc@{}}
\toprule
Vehicle & Signal & Cen & Ent & F20 & F80 & Low & Mid & High \\
\midrule
\multirow{3}{*}{AV-1}
& Speed & 0.509 & 0.767 & 0.829 & 0.414 & 0.591 & 0.165 & 0.313 \\
& Acc   & 0.589 & 0.841 & 0.427 & 0.569 & 0.684 & 0.562 & 0.573 \\
& Jerk  & 0.671 & 0.702 & 0.565 & 0.685 & 0.653 & 0.655 & 0.707 \\
\midrule
\multirow{3}{*}{AV-2}
& Speed & 0.613 & 0.801 & 0.709 & 0.620 & 0.540 & 0.342 & 0.382 \\
& Acc   & 0.701 & 0.771 & 0.621 & 0.627 & 0.629 & 0.689 & 0.593 \\
& Jerk  & 0.771 & 0.780 & 0.632 & 0.751 & 0.769 & 0.757 & 0.722 \\
\midrule
\multirow{3}{*}{AV-3}
& Speed & 0.873 & 0.908 & 0.923 & 0.865 & 0.604 & 0.609 & 0.399 \\
& Acc   & 0.695 & 0.752 & 0.873 & 0.635 & 0.716 & 0.660 & 0.709 \\
& Jerk  & 0.611 & 0.648 & 0.544 & 0.630 & 0.581 & 0.578 & 0.632 \\
\midrule
\multirow{3}{*}{AV-4}
& Speed & 0.752 & 0.885 & 0.818 & 0.689 & 0.586 & 0.334 & 0.575 \\
& Acc   & 0.634 & 0.805 & 0.607 & 0.551 & 0.595 & 0.598 & 0.533 \\
& Jerk  & 0.700 & 0.730 & 0.525 & 0.691 & 0.707 & 0.701 & 0.798 \\
\bottomrule
\end{tabular*}
\end{table}

\begin{table}[!htb]
\centering
\scriptsize
\renewcommand{\arraystretch}{1.05}
\caption{Complete $S^{W}$ values for the medium-density condition across PAV platforms, kinematic signals, and frequency-domain features.}
\label{tab:appendix_w_density_medium}
\begin{tabular*}{\textwidth}{@{\extracolsep{\fill}}llccccccc@{}}
\toprule
Vehicle & Signal & Cen & Ent & F20 & F80 & Low & Mid & High \\
\midrule
\multirow{3}{*}{AV-1}
& Speed & 0.222 & 0.577 & 0.753 & 0.135 & 0.693 & 0.005 & 0.181 \\
& Acc   & 0.326 & 0.753 & 0.149 & 0.401 & 0.407 & 0.326 & 0.413 \\
& Jerk  & 0.665 & 0.658 & 0.559 & 0.724 & 0.658 & 0.665 & 0.675 \\
\midrule
\multirow{3}{*}{AV-2}
& Speed & 0.254 & 0.621 & 0.510 & 0.242 & 0.380 & 0.082 & 0.460 \\
& Acc   & 0.414 & 0.688 & 0.294 & 0.471 & 0.313 & 0.447 & 0.328 \\
& Jerk  & 0.781 & 0.765 & 0.616 & 0.820 & 0.836 & 0.844 & 0.605 \\
\midrule
\multirow{3}{*}{AV-3}
& Speed & 0.784 & 0.806 & 0.876 & 0.765 & 0.859 & 0.366 & 0.298 \\
& Acc   & 0.480 & 0.579 & 0.693 & 0.518 & 0.636 & 0.487 & 0.656 \\
& Jerk  & 0.563 & 0.585 & 0.550 & 0.603 & 0.518 & 0.526 & 0.604 \\
\midrule
\multirow{3}{*}{AV-4}
& Speed & 0.455 & 0.734 & 0.641 & 0.383 & 0.568 & 0.159 & 0.843 \\
& Acc   & 0.265 & 0.559 & 0.379 & 0.256 & 0.257 & 0.238 & 0.240 \\
& Jerk  & 0.550 & 0.780 & 0.349 & 0.651 & 0.586 & 0.597 & 0.730 \\
\bottomrule
\end{tabular*}
\end{table}

\begin{table}[!htb]
\centering
\scriptsize
\renewcommand{\arraystretch}{1.05}
\caption{Complete $S^{\mathrm{KDE}}$ values for the high-density condition across PAV platforms, kinematic signals, and frequency-domain features.}
\label{tab:appendix_kde_density_high}
\begin{tabular*}{\textwidth}{@{\extracolsep{\fill}}llccccccc@{}}
\toprule
Vehicle & Signal & Cen & Ent & F20 & F80 & Low & Mid & High \\
\midrule
\multirow{3}{*}{AV-1}
& Speed & 0.717 & 0.746 & 0.846 & 0.650 & 0.557 & 0.534 & 0.382 \\
& Acc   & 0.696 & 0.786 & 0.702 & 0.610 & 0.575 & 0.661 & 0.670 \\
& Jerk  & 0.676 & 0.759 & 0.645 & 0.694 & 0.751 & 0.736 & 0.781 \\
\midrule
\multirow{3}{*}{AV-2}
& Speed & 0.651 & 0.759 & 0.802 & 0.629 & 0.626 & 0.323 & 0.559 \\
& Acc   & 0.687 & 0.615 & 0.604 & 0.629 & 0.792 & 0.606 & 0.659 \\
& Jerk  & 0.703 & 0.795 & 0.647 & 0.694 & 0.766 & 0.833 & 0.501 \\
\midrule
\multirow{3}{*}{AV-3}
& Speed & 0.818 & 0.746 & 0.919 & 0.717 & 0.667 & 0.631 & 0.464 \\
& Acc   & 0.706 & 0.656 & 0.599 & 0.688 & 0.775 & 0.682 & 0.652 \\
& Jerk  & 0.696 & 0.668 & 0.655 & 0.656 & 0.653 & 0.663 & 0.628 \\
\midrule
\multirow{3}{*}{AV-4}
& Speed & 0.695 & 0.701 & 0.807 & 0.595 & 0.560 & 0.481 & 0.620 \\
& Acc   & 0.812 & 0.937 & 0.712 & 0.712 & 0.641 & 0.721 & 0.584 \\
& Jerk  & 0.690 & 0.863 & 0.596 & 0.741 & 0.729 & 0.727 & 0.682 \\
\bottomrule
\end{tabular*}
\end{table}

\begin{table}[!htb]
\centering
\scriptsize
\renewcommand{\arraystretch}{1.05}
\caption{Complete $S^{W}$ values for the high-density condition across PAV platforms, kinematic signals, and frequency-domain features.}
\label{tab:appendix_w_density_high}
\begin{tabular*}{\textwidth}{@{\extracolsep{\fill}}llccccccc@{}}
\toprule
Vehicle & Signal & Cen & Ent & F20 & F80 & Low & Mid & High \\
\midrule
\multirow{3}{*}{AV-1}
& Speed & 0.520 & 0.476 & 0.639 & 0.458 & 0.550 & 0.829 & 0.486 \\
& Acc   & 0.654 & 0.621 & 0.706 & 0.526 & 0.486 & 0.689 & 0.642 \\
& Jerk  & 0.604 & 0.720 & 0.624 & 0.688 & 0.788 & 0.766 & 0.716 \\
\midrule
\multirow{3}{*}{AV-2}
& Speed & 0.482 & 0.435 & 0.670 & 0.463 & 0.638 & 0.720 & 0.789 \\
& Acc   & 0.643 & 0.371 & 0.703 & 0.614 & 0.777 & 0.673 & 0.601 \\
& Jerk  & 0.571 & 0.785 & 0.585 & 0.628 & 0.751 & 0.819 & 0.171 \\
\midrule
\multirow{3}{*}{AV-3}
& Speed & 0.710 & 0.455 & 0.850 & 0.562 & 0.895 & 0.887 & 0.395 \\
& Acc   & 0.700 & 0.465 & 0.687 & 0.674 & 0.732 & 0.726 & 0.612 \\
& Jerk  & 0.576 & 0.467 & 0.635 & 0.553 & 0.533 & 0.534 & 0.477 \\
\midrule
\multirow{3}{*}{AV-4}
& Speed & 0.443 & 0.359 & 0.558 & 0.376 & 0.453 & 0.809 & 0.847 \\
& Acc   & 0.726 & 0.884 & 0.679 & 0.615 & 0.481 & 0.725 & 0.383 \\
& Jerk  & 0.598 & 0.865 & 0.473 & 0.703 & 0.693 & 0.731 & 0.470 \\
\bottomrule
\end{tabular*}
\end{table}

\clearpage
\section{Complete KDE Distributions}
\label{app:complete_kde}
\setcounter{figure}{0}

\begin{figure}[!htbp]
    \centering
    \includegraphics[width=\textwidth]{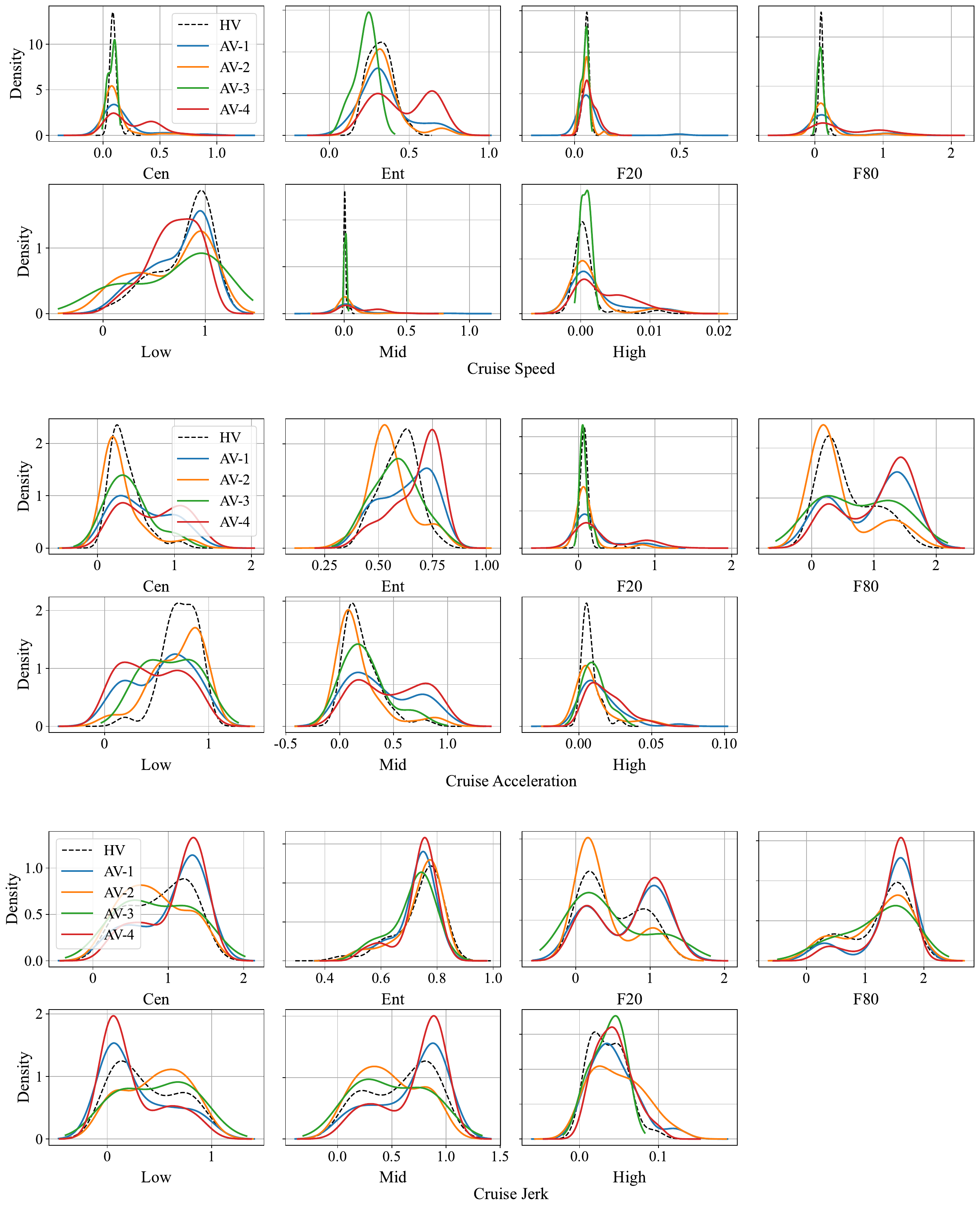}
    \caption{Complete KDE distributions of the seven spectral features for speed, acceleration, and jerk under free-flow cruising conditions.}
    \label{fig:app_kde_cruise}
\end{figure}

\begin{figure}[!htbp]
    \centering
    \includegraphics[width=\textwidth]{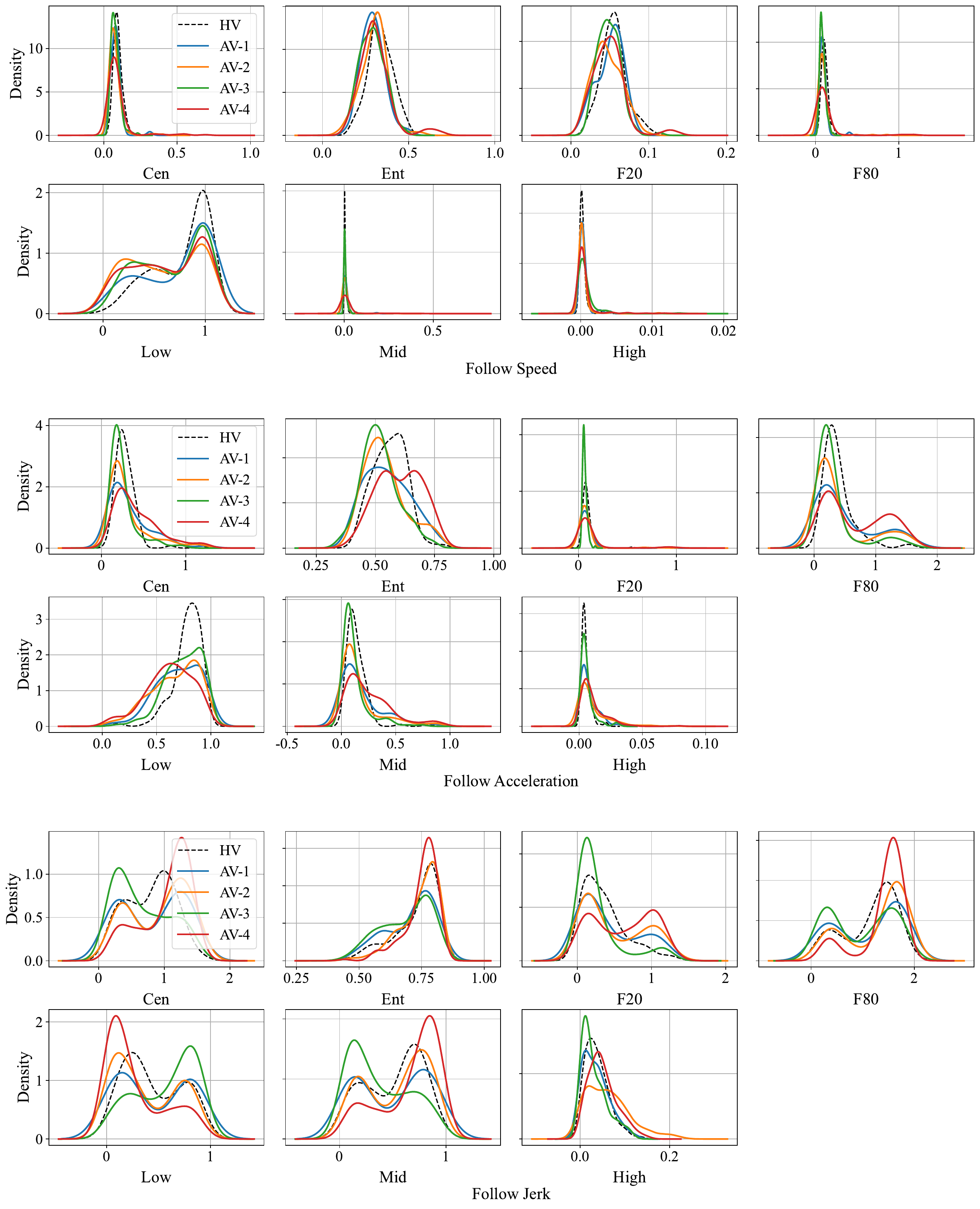}
    \caption{Complete KDE distributions of the seven spectral features for speed, acceleration, and jerk under car-following conditions.}
    \label{fig:app_kde_follow}
\end{figure}

\begin{figure}[!htbp]
    \centering
    \includegraphics[width=\textwidth]{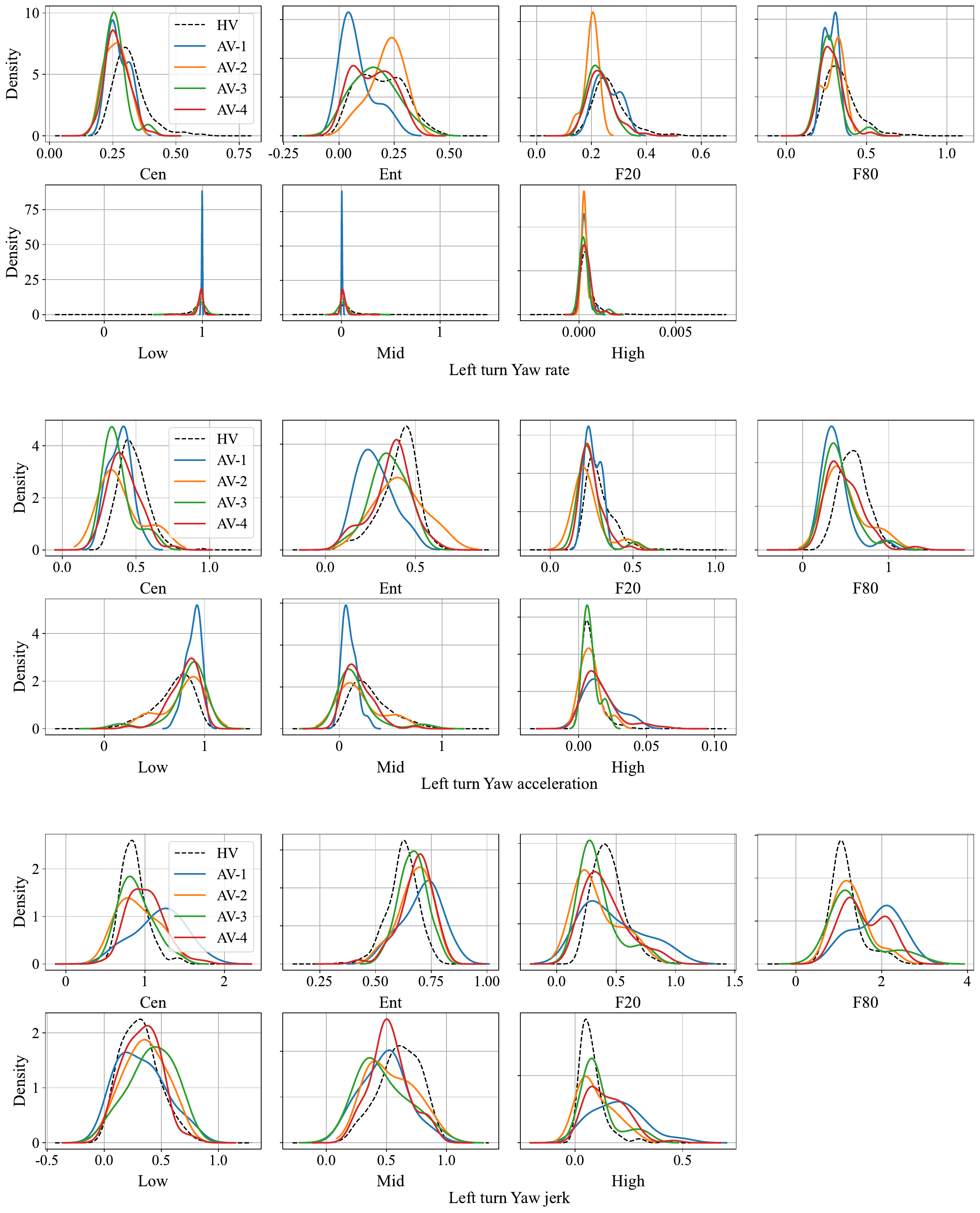}
    \caption{Complete KDE distributions of the seven spectral features for yaw rate, yaw acceleration, and yaw jerk under left-turn conditions.}
    \label{fig:app_kde_left}
\end{figure}

\begin{figure}[!htbp]
    \centering
    \includegraphics[width=\textwidth]{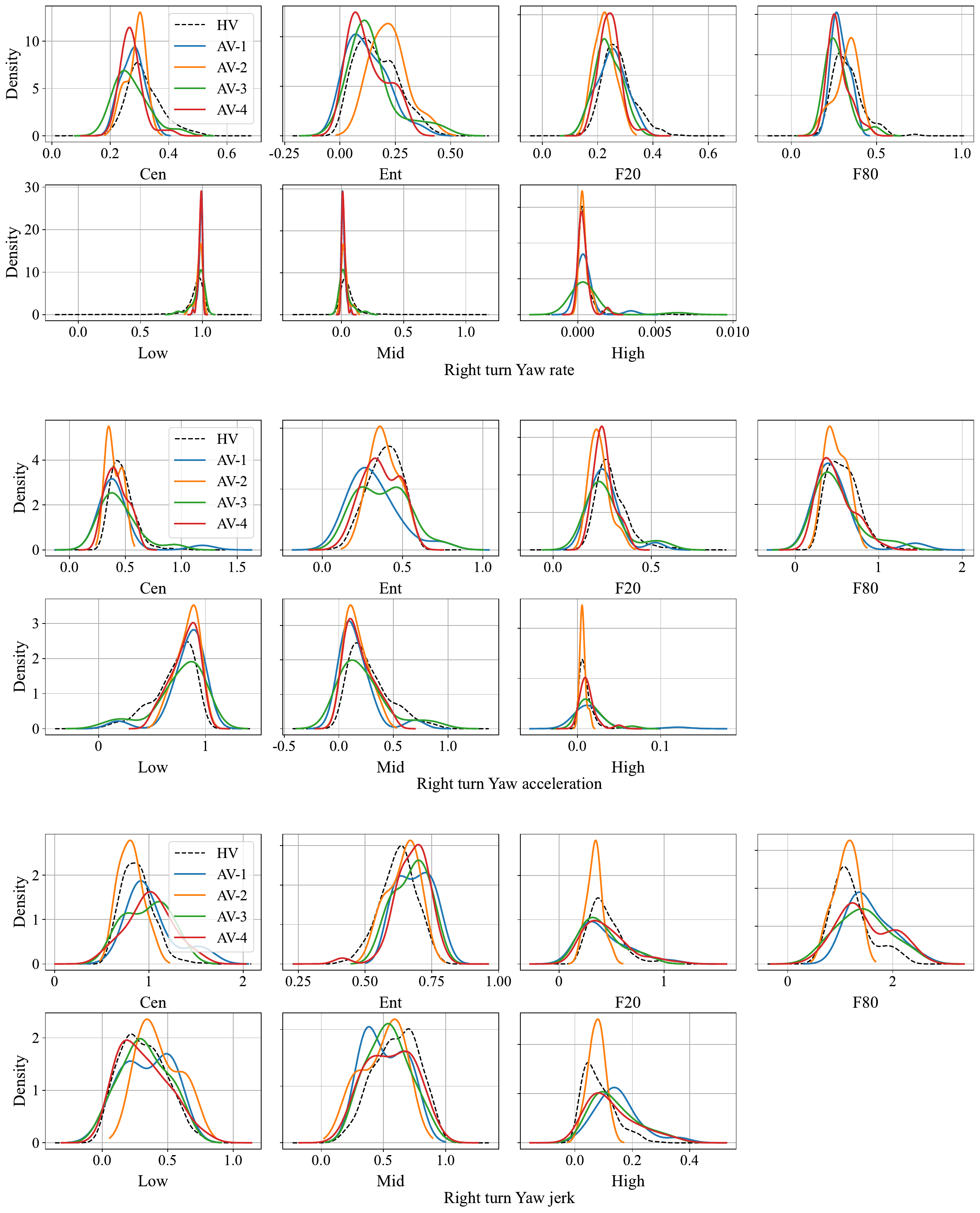}
    \caption{Complete KDE distributions of the seven spectral features for yaw rate, yaw acceleration, and yaw jerk under right-turn conditions.}
    \label{fig:app_kde_right}
\end{figure}

\begin{figure}[!htbp]
    \centering
    \includegraphics[width=\textwidth]{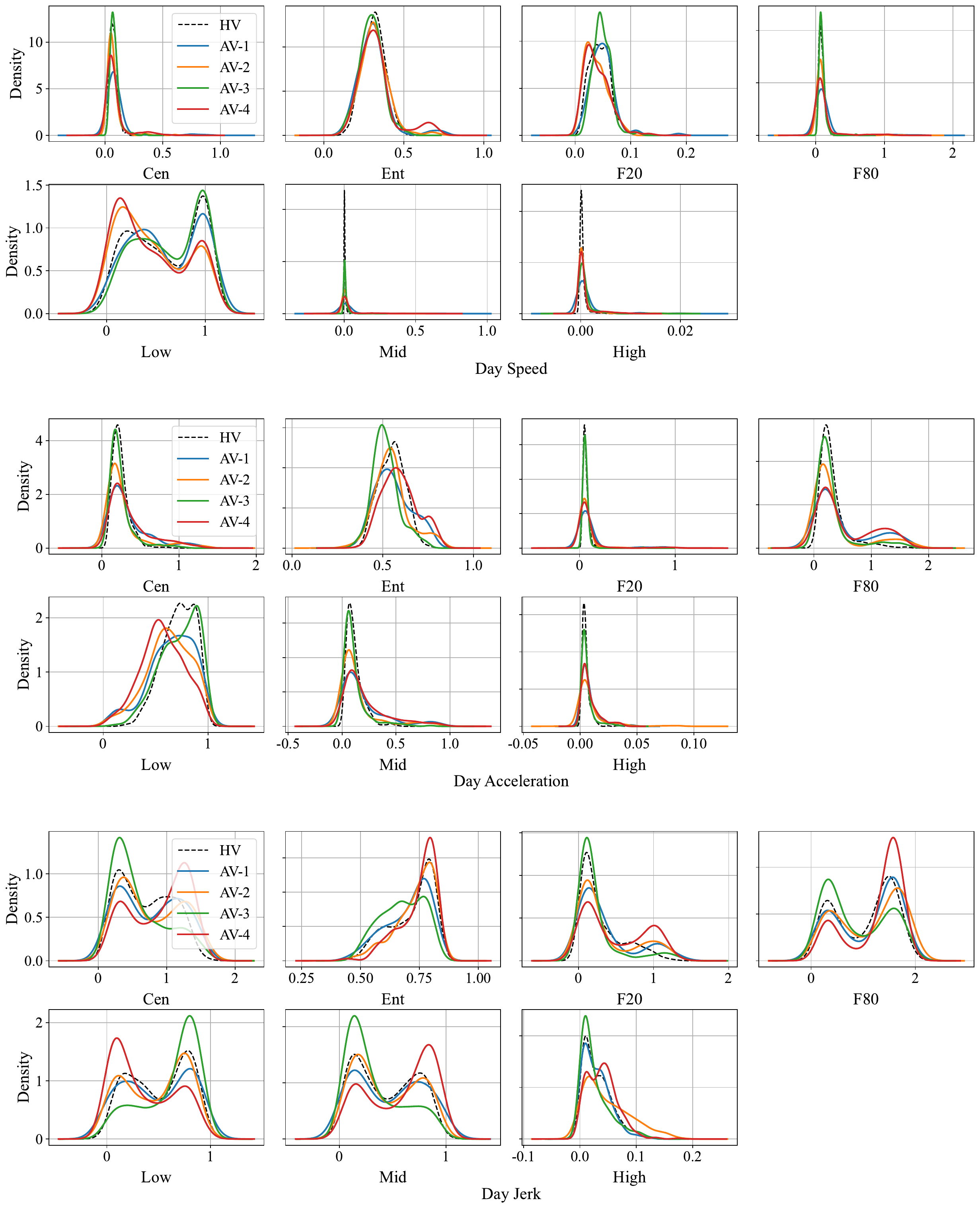}
    \caption{Complete KDE distributions of the seven spectral features for speed, acceleration, and jerk under daytime conditions.}
    \label{fig:app_kde_day}
\end{figure}

\begin{figure}[!htbp]
    \centering
    \includegraphics[width=\textwidth]{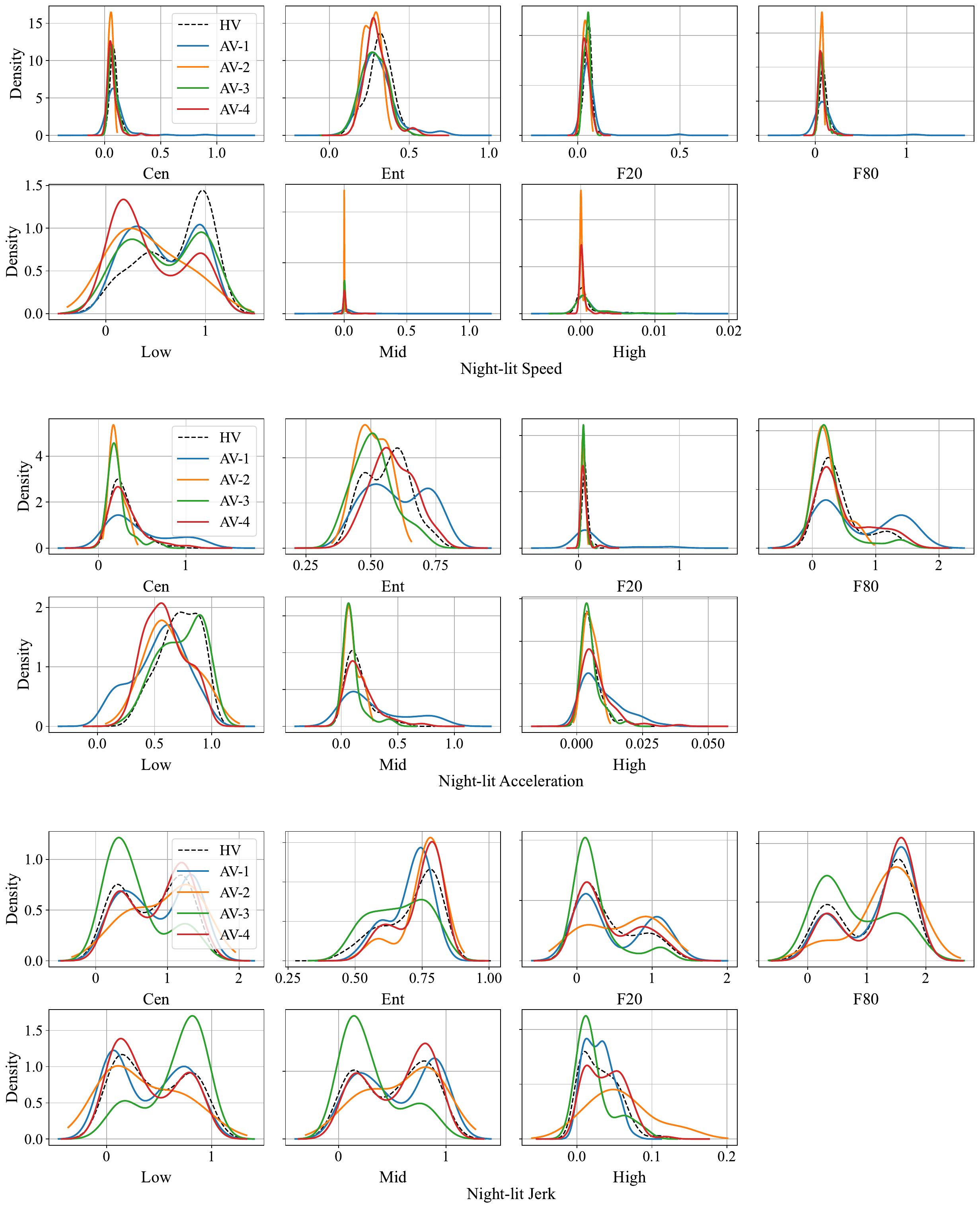}
    \caption{Complete KDE distributions of the seven spectral features for speed, acceleration, and jerk under night-lit conditions.}
    \label{fig:app_kde_night_lit}
\end{figure}

\begin{figure}[!htbp]
    \centering
    \includegraphics[width=\textwidth]{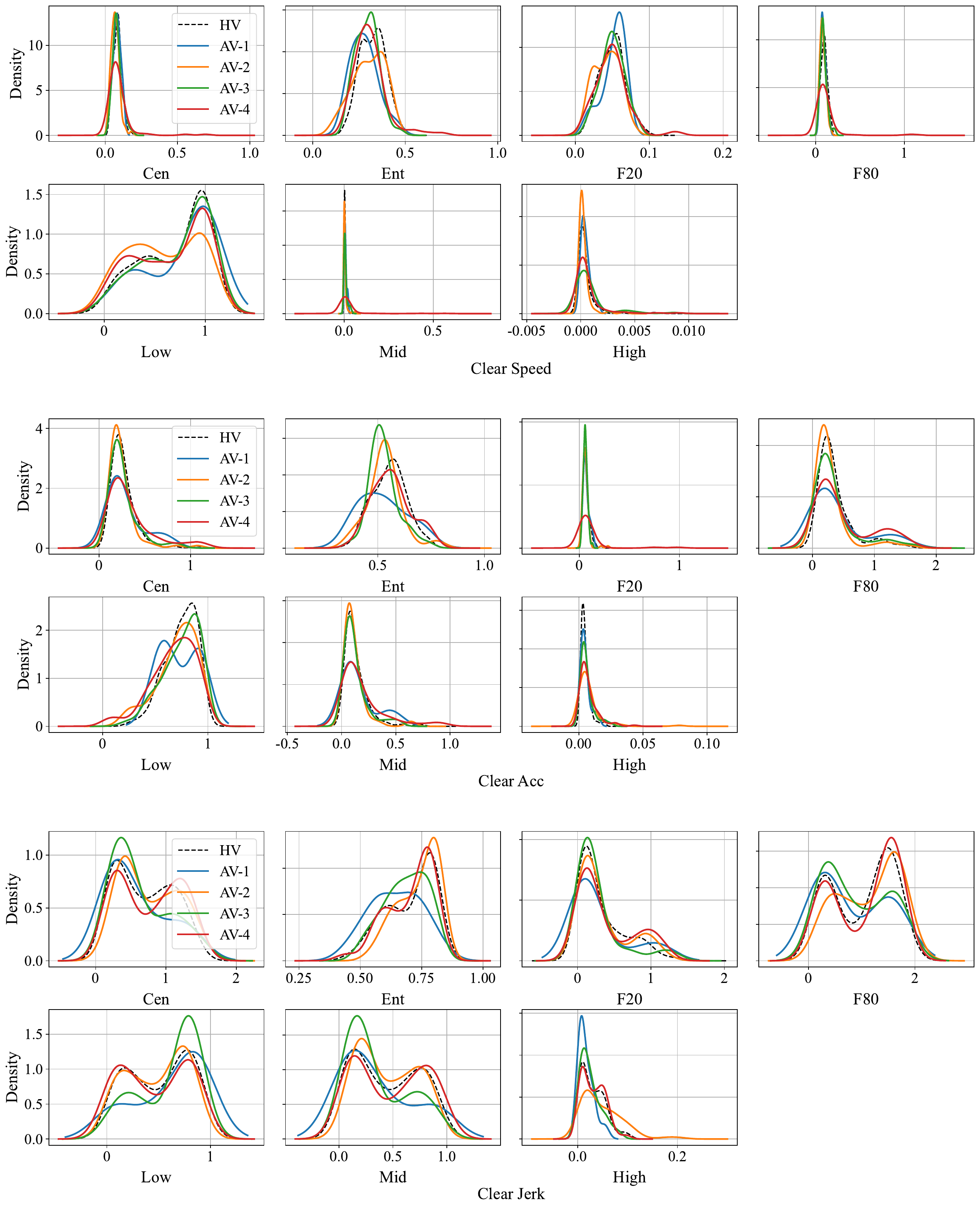}
    \caption{Complete KDE distributions of the seven spectral features for speed, acceleration, and jerk under clear-weather conditions.}
    \label{fig:app_kde_clear}
\end{figure}

\begin{figure}[!htbp]
    \centering
    \includegraphics[width=\textwidth]{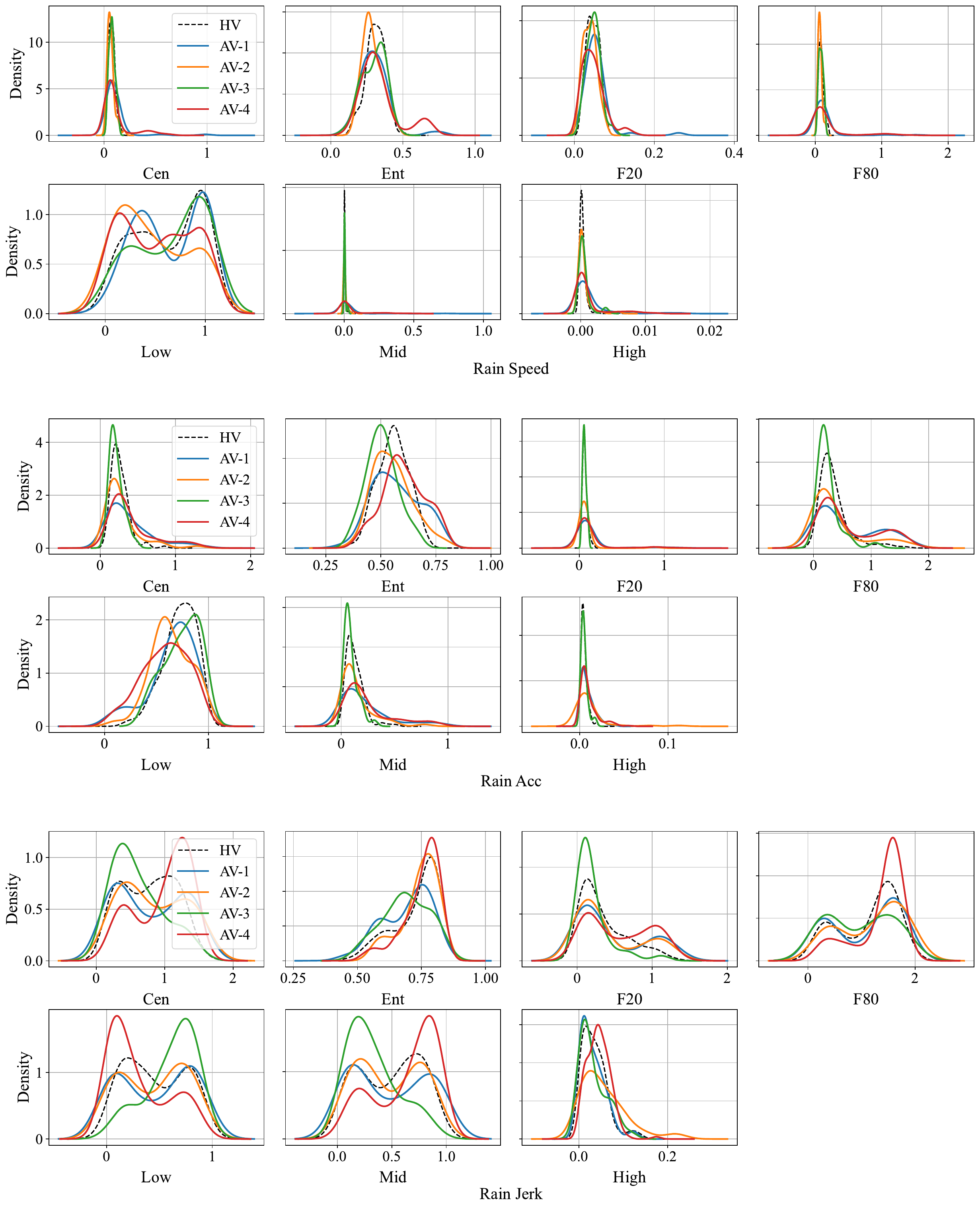}
    \caption{Complete KDE distributions of the seven spectral features for speed, acceleration, and jerk under rainy conditions.}
    \label{fig:app_kde_rain}
\end{figure}

\begin{figure}[!htbp]
    \centering
    \includegraphics[width=\textwidth]{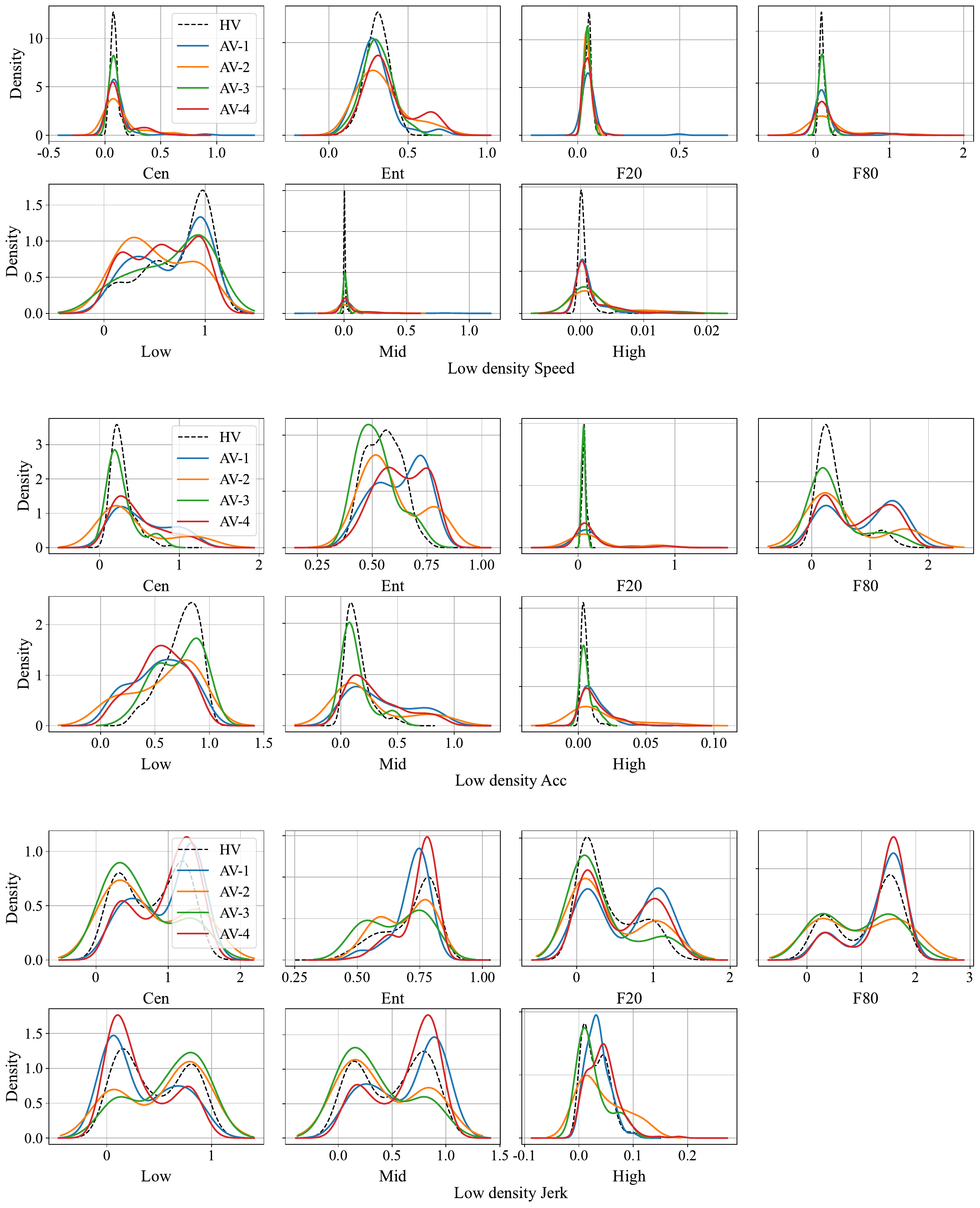}
    \caption{Complete KDE distributions of the seven spectral features for speed, acceleration, and jerk under low vehicle-density conditions.}
    \label{fig:app_kde_density_low}
\end{figure}

\begin{figure}[!htbp]
    \centering
    \includegraphics[width=\textwidth]{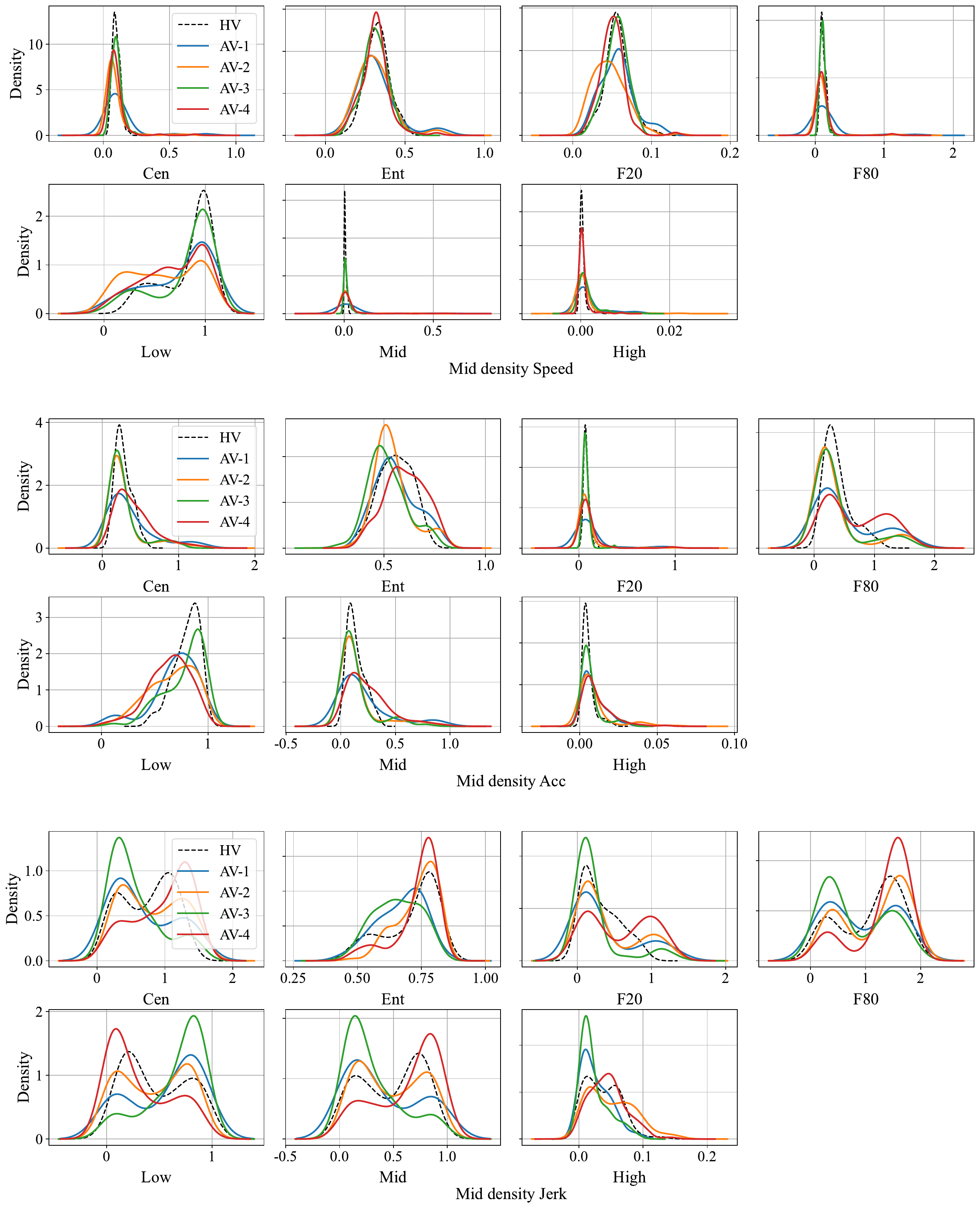}
    \caption{Complete KDE distributions of the seven spectral features for speed, acceleration, and jerk under medium vehicle-density conditions.}
    \label{fig:app_kde_density_medium}
\end{figure}

\begin{figure}[!htbp]
    \centering
    \includegraphics[width=\textwidth]{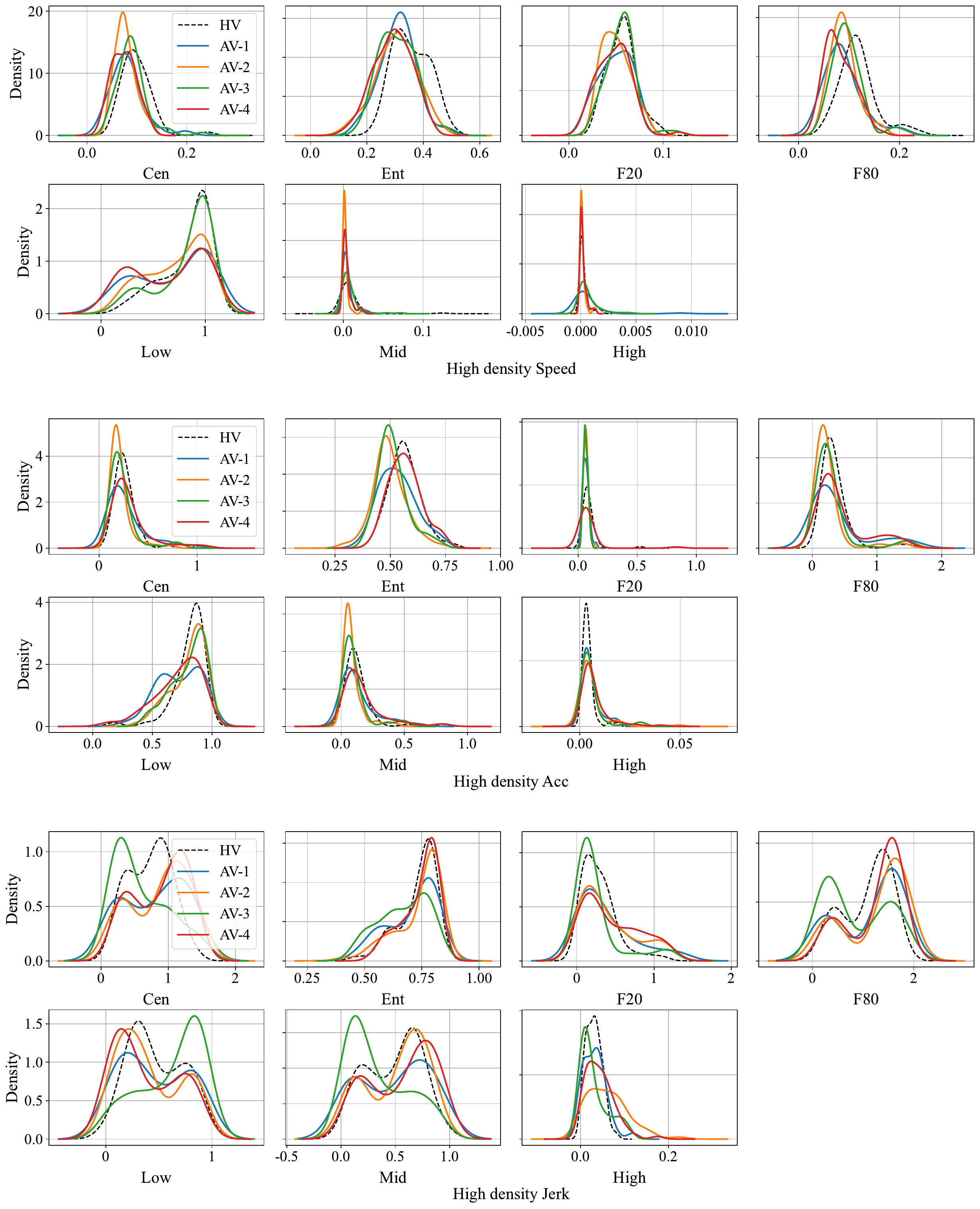}
    \caption{Complete KDE distributions of the seven spectral features for speed, acceleration, and jerk under high vehicle-density conditions.}
    \label{fig:app_kde_density_high}
\end{figure}

\end{document}